\documentclass{article}

\usepackage[preprint]{neurips_2024}

\usepackage{natbib}
\usepackage[utf8]{inputenc} 
\usepackage[T1]{fontenc}    
\usepackage[hypertexnames=false]{hyperref}       
\usepackage{url}            
\usepackage{booktabs}       
\usepackage{amsfonts}       
\usepackage{nicefrac}       
\usepackage{microtype}      
\usepackage{xcolor}         

\usepackage{graphicx} 
\usepackage[english]{babel}

\usepackage{amsmath}
\usepackage{amssymb}
\usepackage{amsthm}
\usepackage{bbm} 

\usepackage[normalem]{ulem} 

\usepackage{amsbsy}
\usepackage{tikz}
\usetikzlibrary{positioning}
\usepackage{stmaryrd}
\usepackage{color}
\usepackage{colortbl} 
\usepackage{makeidx}
\usepackage[shortlabels]{enumitem} 

\usepackage{caption}
\usepackage{subcaption}
\usepackage{arydshln}  
\usepackage{makecell} 
\usepackage{multirow}  

\usepackage{algorithm}
\usepackage{algpseudocode}

\usepackage{cleveref}
\crefname{enumi}{part}{parts}
\usepackage{comment}

\usepackage{geometry}
\usepackage{graphicx}
\usepackage{wrapfig}
\usepackage{mdframed} 

\usepackage{ragged2e}

\newtheorem{thm}{Theorem}[section]
\crefname{thm}{Theorem}{Theorems}
\newtheorem{lemma}{Lemma}[section]
\crefname{lemma}{Lemma}{Lemmas}
\newtheorem{cond}{Assumption}[section]
\crefname{cond}{Assumption}{Assumptions}

\crefname{mech}{Mechanism}{Mechanisms}

\crefname{algorithm}{Algorithm}{Algorithms}

\crefname{table}{Table}{Tables}
\crefname{figure}{Figure}{Figures}

\theoremstyle{remark}
\newtheorem{rem}{Remark}[section]
\crefname{rem}{remark}{remarks}

\theoremstyle{definition}
\newtheorem{defi}{Definition}[section]
\newtheorem{bsp}{Example}[section]

\newcommand{\argmin}[1]{\underset{#1}{\textnormal{argmin }}}
\newcommand{\E}{\mathbb{E}}
\newcommand{\p}{\mathbb{P}}

\DeclareMathOperator{\Unif}{Unif}

\DeclareMathOperator{\Lap}{Lap}

\title{Uncertainty quantification by block bootstrap for differentially private stochastic gradient descent}
\author{%
	Holger Dette \\
	Chair of Stochastics\\
	Ruhr University Bochum\\
	Germany, Bochum\\
	\texttt{holger.dette@ruhr-uni-bochum.de} \\
	\And
	Carina Graw \\
	Chair of Stochastics\\
	Ruhr University Bochum\\
	Germany, Bochum\\
	\texttt{carina.graw@ruhr-uni-bochum.de} \\
}
\date{}

\begin{document}
	
	\maketitle
	\begin{abstract}
		\noindent Stochastic Gradient Descent (SGD) is a widely used tool in machine learning. 
		In the context of Differential Privacy (DP), SGD has been well studied in the last years in which the focus is mainly on convergence rates and privacy guarantees. 
		While in the non private case, uncertainty quantification (UQ) for  SGD by bootstrap has been addressed by several authors, these  procedures cannot be transferred to differential privacy due to multiple queries to the private data. 
		In this paper, we propose a novel block bootstrap for SGD under local differential privacy that is computationally tractable and does not require an adjustment of the privacy budget. 
		The method can be easily  implemented and is applicable to a broad class of estimation problems.
		We prove the validity of our approach and illustrate its finite sample properties by means of a simulation study. 
		As a by-product, the new method also provides a simple alternative numerical tool for UQ for non-private SGD.
	\end{abstract}

	\section{Introduction}
	In times where data is collected almost everywhere and anytime, privacy is an important issue that has to be addressed in any data analysis. 
	In the past, successful de-anonymisation attacks have been performed 
	\cite[see, for example, ][]{narayanan2006break,gambs2014anonymization,eshun2022two} leaking private, possible sensitive data of individuals.
	Differential Privacy (DP) is a framework that has been introduced by \citet{dwork2006} to protect an individuals data against such attacks while still learning something about the population.
	Nowadays, Differential Privacy has become a state-of-the-art framework which has been implemented by the US Census and large companies like Apple, Microsoft and Google \cite[see][]{ding2017collecting,abowd2018us}.\\
	The key idea of Differential Privacy is to randomize the data or a data dependent statistic. 
	This additional randomness guarantees that the exchange of one individual merely changes the distribution of the randomized output.
	In the last decade, numerous differentially private algorithms have been developed, see for example \citet{wang2020comprehensive,xiong2020comprehensive,yang2023local} and the references therein. 
	In many cases differentially private algorithms are based on already known and well studied procedures
	such as empirical risk minimisation, where either the objective function (objective pertubation) or the minimizer of that function (output perturbation) is privatized, or statistical testing,  where the statistic is privatized, see for example \citet{chaudhuri2011differentially,vu2009differential}. 
	Two different concepts of differential privacy are distinguished in the literature: 
	the first one,  called (global) {\it Differential Privacy} (DP) assumes 
	the presence of a trusted curator, who maintains and performs computations on the data and ensures that a published statistic satisfies a certain privacy guarantee. 
	The second one, which is considered in this paper, does not make this assumption and is called {\it local Differential Privacy} (LDP). 
	Here it is required that the data is collected in a privacy preserving way.
	\smallskip\\
	Besides  privacy there is  a second issue which has to be addressed in the analysis of massive data, namely the problem of  computational complexity.  A common and widely used tool when dealing with  large-scale and complex optimization problems in  big data analysis 
	is {\it Stochastic Gradient Descent} (SGD), which is already introduced in the early work of  \citet{robbins1951stochastic}.
	This procedure computes data based, iteratively and in an online fashion
	an estimate of the minimizer 
	\begin{align} \label{det0}
		\theta_\star = \argmin{\theta}  L(\theta)
	\end{align} 
	of a function  $L$.
	The convergence properties of adaptive algorithms, such as SGD, towards $\theta_\star$ are well studied, see for example
	the monograph \citet{benveniste1990adaptive} and the references therein.
	\citet{mertikopoulos2020almost} investigate SGD in case of non convex functions and show convergence to a local minimizer.\\
	There also exists a large amount of work on statistical inference based on SGD estimation starting with the seminal papers of \citet{ruppert1988efficient} and \citet{Polyak1992}, who establish asymptotic normality of the averaged SGD estimator for convex functions.
	The covariance matrix of the asymptotic distribution has a sandwich form, say $G^{-1}SG^{-1}$, and  \citet{chen2020} propose two different approaches to estimate this matrix. An online version for one of these methods  is proposed in  \citet{Zhu2021}, while \citet{yu2021analysis} generalize these results for SGD with constant stepsize
	and  non convex loss functions.  
	Recent literature on  UQ for  SDG includes  \citet{su2018uncertainty}, who  propose  iterative sample splitting of the path of SGD,  
	\citet{li2018statistical}, who consider SGD with constant stepsize splitting the  whole path into segments,  and   \citet{chee2023plus}, who construct a simple confidence interval for the last iterate of SGD.
	\\
	\citet{song2013} investigate SGD under DP and demonstrate that mini-batch SGD  has a much better accuracy than estimators obtained with batch size $1$. 
	However, mini-batch SGD with a  batch size larger than $1$ only guarantees global DP  and therefore requires the presence of a trusted curator.
	Among other approaches for empirical risk minimization, \citet{bassily2014} derive  a utility bound in terms of the excess risk for an $(\epsilon,\delta)$-DP SGD and 
	\citet{avella2023differentially} investigate statistical inference for noisy gradient descent under Gaussian DP.
	\\
	On the other hand, statistcial inference for SGD under LDP is far less explored. 
	\citet{duchi18minimax} show the  minimax optimality of the LDP SGD median but do not provide confidence intervals for inference.
	The  method of  \cite{chen2020}  is not always applicable  (for example, for quantiles) and requires an additional private estimate of the matrix $G$
	in the asymptotic covariance matrix 
	of the SDG estimate.  Recently, 
	\citet{liu2023online} propose a self-normalizing approach to obtain 
	distribution free locally differentially private confidence intervals 
	for quantiles.
	However, this method is tailored to a specific model and it is well known that the nice property of distribution free inference by  self-normalization  comes with the price of a loss in statistical efficiency \cite[see][for an example in the context of testing]{DetteGoesmann}.
	\\
	A natural alternative  is the application of bootstrap, and bootstrap for SGD   in the non-private case has been studied  by \citet{fang2018,zhong2023}.  These authors  propose  a multiplier approach where  SGD is executed $B$ times  multiplying the  updates by  non-negative random variables.
	However, to ensure 
	$\epsilon$-LDP, the privacy budget must be split over all $B$ bootstrap versions, leading to large noise levels and high inaccuracy of the resulting estimators.
	There is also ongoing research on parametric Bootstrap under DP, \cite[see][]{ferrando2022parametric, fu2023differentially, wangdebiased} who assume that the parameter of interest characterizes the distribution of the data, which is not necessarily the case for SGD.  
	Finally we mention \citet{wang2022}, who introduce a bootstrap procedure under Gaussian DP, where each bootstrap estimator is privately computed on a random subset.
	
	\textbf{Our Contribution: } In this paper we propose a computational tractable method for statistical inference using SGD under LDP. 
	Our approach provides a universally applicable and statistically valid method for uncertainty quantification for SDG and LDP-SGD.  
	It is based on an application of the  block bootstrap to the mean of the iterates obtained by SGD. 
	The (multiplier)  block bootstrap is  a common principle in time series analysis \cite[see][]{lahiri2003}. 
	However, in contrast to this field, where the dependencies in the sequence of observations are quickly decreasing with an increasing lag, the statistical  analysis of such an approach for SGD is very challenging. 
	More precisely, SGD produces a sequence of highly dependent iterates, which makes the application of classical concepts from time series analysis such as mixing \cite[see][]{bradley2005} or physical dependence \cite[see][]{Wu2005} impossible. 
	Instead, we use a different technique and   show the  consistency of the proposed multiplier block bootstrap for the LDP-SGD  
	under appropriate conditions on the block length and  the learning rate. 
	As a by-product our results also provide a new method of UQ for SGD and for mini-batch SGD in the non-private case.
	
	{\bf Notations:} $\lVert \cdot \rVert$ denotes a norm on $\mathbb{R}^d$ if not further specified. $\overset{d}{\to}$ denotes weak convergence and $\overset{\p}{\to}$ denotes convergence in probability.

	\section{Preliminaries and Background} 
	
	{\bf  Stochastic Gradient Descent (SGD).}
	Define $L(\theta)=\E[\ell(\theta,X)]$, where $X$ is a $\mathbb{X}$-valued random variable  and $\ell:\mathbb{R}^d\times \mathbb{X}\to \mathbb{R}$ is a loss function. 
	We consider the optimization problem \eqref{det0}.
	If  $L$ is differentiable and convex, then this minimization problem is equivalent to solving
	\begin{equation*}
		R(\theta) :=  \nabla L(\theta)=0 ,  
	\end{equation*}
	where $R= \nabla L$ is the gradient of $L$.
	SGD computes an estimator $\bar{\theta}$ for $\theta_\star$ using noisy observations $g(X_i, \cdot )$ of the gradient $R(\cdot )$ in an iterative way. Note that $g$ does not need to be differentiable with respect to $\theta$. The iterations of SGD are defined by 
	\begin{align} \label{det4}
		\theta_i&=\theta_{i-1} - \eta_i g(X_i,\theta_{i-1}) 
		= \theta_{i-1} - \eta_i (R(\theta_{i-1}) +\xi_i) 
	\end{align}
	where $\eta_i=ci^{-\gamma}$ with parameters $c>0$ and $\gamma \in (0.5,1) $ is the learning rate and  the quantity  $\xi_i= g(X_i,\theta_{i-1})-R(\theta_{i-1})$ is called {\it  measurement error} in the $i$th iteration
	(note that we do not reflect the dependence on $\theta_{i-1} $ and $X_i$ in the definition of $\xi_i$).
	The estimator of $\theta_\star$ is finally obtained as the average of the iterates, that is  
	\begin{align} \label{det5}
		\bar{\theta}&=\frac{1}{n}\sum_{i=1}^n\theta_i ~. 
	\end{align}
	Note that the first representation in \eqref{det4} can be directly used for implementation, while the second is helpful for proving probabilistic properties of SGD. 
	For example  \citet{Polyak1992} show that, under appropriate conditions,
	$\sqrt{n}(\bar{\theta}-\theta_\star)$ is asymptotically normal distributed 
	where the covariance matrix $\Sigma$ of the limit distribution depends on the optimization problem and the variance of the measurement errors $\xi_t$.
	
	The use of only  one observation $g(X_i,\theta_{i-1})$  in 
	SGD yields a relatively large measurement error in each  iteration
	which might result in  inaccurate estimation of the gradient $R(\theta_{i-1})$.
	Therefore, several authors have  proposed a variant of SGD, called
	mini-batch SGD, where $s\geq 1$ observations $g(X_{i_1},\tilde{\theta}_{i-1}),\ldots,g(X_{i_s},\tilde{\theta}_{i-1})$ are used for the update.
	An estimator for $R(\tilde \theta_{i-1})$ is then given by the mean of this observations, yielding the recursion
	\begin{align} \label{det6}
		\tilde{\theta}_{i}&=\tilde{\theta}_{i-1} - \eta_i \frac{1}{s}\sum_{j=1}^s g(X_{i_j},\tilde{\theta}_{i-1})
		= \tilde{\theta}_{i-1} - \eta_i (R(\tilde{\theta}_{i-1}) +\tilde{\xi}_i)
	\end{align} 
	where the measurement error is given by $\tilde{\xi}_i= \frac{1}{s}\sum_{j=1}^s g(X_{i_j},\tilde{\theta}_{i-1})-R(\tilde{\theta}_{i-1})$,
	see for example \citet{Shamir2011,Khirirat2017}.
	
	\medskip
	
	\noindent 
	{\bf  Differential Privacy (DP).} 
	The idea of differential privacy is that one individual should not change the outcome of an algorithm largely, i.e. changing one data point should not alter the result too much. 
	Let $(\mathbb{X},\mathcal{X})$ be a measurable space. For $n\in\mathbb{N}$, $x\in\mathbb{X}^n$ will be called a data base containing $n$ data points.
	Two data bases $x,x'\in\mathbb{X}^n$ are called  neighbouring if they only differ in one data point and are called disjoint if all data points are different.
	How much a single data point is allowed to change the outcome is captured in a privacy parameter $\epsilon$. 
	Let $(\mathbb{Y},\mathcal{Y})$ be a measurable space. A randomized algorithm $\mathcal{A}: \mathbb{X}^n \to \mathbb{Y}$ 
	maps $x\in\mathbb{X}^n$ onto a random variable $\mathcal{A}(x)$ with values in $\mathbb{Y}$.  
	Here we will consider either subspaces of $\mathbb{R}^d$ equipped with the Borel sets or finite sets equipped with their power set.

	\begin{defi}
		Let $\epsilon >0$.
		A randomized algorithm $\mathcal{A}: \mathbb{X}^n\to \mathbb{Y}$ is $\epsilon$-{\it differentially private} (dp) if for all neighbouring $x,x'\in \mathbb{X}^n$ and all sets $S\in \mathcal{Y}$
		\begin{equation} \label{dp eq}
			P(\mathcal{A}(x)\in S)\leq e^\epsilon P(\mathcal{A}(x')\in S).
		\end{equation}
		$\mathcal{A}$ is also called a $\epsilon$-dp mechanism and is said to preserve DP.
		If $\mathcal{A}$ is restricted to $n=1$, i.e. $x$ and $x'$ contain only one data point,  and \eqref{dp eq} holds, $\mathcal{A}$ is  $\epsilon$-{\it local differentially private} (ldp). 
	\end{defi}
	In this paper we work under the constraints of local differential privacy (LDP).
	
	The following result is well known \cite[see, for example][]{dwork2014algorithmic}.
	
	\newpage
	\begin{thm} \label{Post}
		~~~
		\begin{enumerate}
			\item {\rm Post processing:} Let $(\mathbb{Z}, \mathcal{Z})$ be a measurable space, $\mathcal{A}: \mathbb{X}^n\to \mathbb{Y}$ be an $\epsilon$-dp mechanism and $f: \mathbb{Y}\to \mathbb{Z}$ be a measurable function. Then $f\circ \mathcal{A}$ is $\epsilon$-dp.
			\item {\rm Sequential composition:}  
			Let $\mathcal{A}_i: \mathbb{X}^n\to \mathbb{Y}$, $i=1,\ldots, k$ be $\epsilon_i$-dp mechanisms. Then $\mathcal{A}: \mathbb{X}^{n}\to \mathbb{Y}^k$ that maps $x\mapsto(\mathcal{A}_1(x),\ldots,\mathcal{A}_k(x))$ is $\sum_{i=1}^k\epsilon_i$-dp.
			\item {\rm Parallel composition:}  
			Let $\mathcal{A}_i: \mathbb{X}^n\to \mathbb{Y}$, $i=1,\ldots, k$ be $\epsilon_i$-dp mechanisms and $x_1,\ldots,x_k \in \mathbb{X}^n$ be disjoint. Then $\mathcal{A}: \mathbb{X}^{n\times k}\to \mathbb{Y}^k$ that maps $(x_1,\ldots,x_k)\mapsto(\mathcal{A}_1(x_1),\ldots,\mathcal{A}_k(x_k))$ is $\max \epsilon_i$-dp. 
		\end{enumerate}   
	\end{thm}
	
	Two well known privacy mechanism
	are the following:
	\begin{bsp}[Laplace Mechanism] \label{Laplace}
		Let $f: \mathbb{X} \to \mathbb{Y}\subset\mathbb{R}$ be a function. Its sensitivity is defined as $\Delta(f)=\sup_{x,x'\in \mathbb{X} } \lVert f(x)-f(x')\rVert$. Assume that $\Delta(f)<\infty$ and let $L\sim \Lap(0,\Delta(f)/\epsilon)$,
		where $\Lap(0,b)$ denotes a centered Laplace distribution 
		with density
		\[
		p(x) =  \frac{1}{2b}\exp\left(-\frac{|x|}{b}\right).
		\]
		Then the randomized algorithm $\mathcal{A}_{Lap}: x \mapsto f(x)+L$ is $\epsilon$-dp.
	\end{bsp}
	\begin{bsp}[Randomized Response]  \label{Randomized Response}
		Let $f:\mathbb{X}\to \{0,1\}$ be a function. Denote $p=\frac{e^\epsilon}{e^\epsilon+1}$ and define for $x \in \mathbb{X}$ a random variable $\mathcal{A}_{RR}(x)$ on $\{0,1\}$ with conditional distribution given $x$ as
		\[\mathcal{A}_{RR}(x) \sim \begin{cases}
			B(p)& \textnormal{, } f(x) = 1\\
			B(1-p)& \textnormal{, } f(x) = 0
		\end{cases},\]
		where $B(p)$ denotes a Bernoulli distribution with success probability $p$.
		Then $ \mathcal{A}_{RR}(x) $ is $\epsilon$-dp.
	\end{bsp}
	
	\medskip
	
	{\bf Local Differential Private  Stochastic Gradient Descent 
		(LDP-SGD). } 
	By \cref{Post} it follows  that SGD is $\epsilon$-ldp if the noisy observations
	$g(X_i,\theta_{i-1})$ in \eqref{det4} are  computed in a way that preserves $\epsilon$-LDP. 
	Let $\mathcal{A}$ be such an $\epsilon$-ldp mechanism for computing $g$. The LDP-SGD updates are then given by
	\begin{align}
		\theta_i^{LDP}=&\theta_{i-1}^{LDP}-\eta_i \mathcal{A}(g(X_i,\theta^{LDP}_{i-1}))
		= \theta_{i-1}^{LDP}-\eta_i \big(R(\theta_{i-1}^{LDP}) + \xi_i^{SGD} + \xi_i^{LDP}\big) \label{DP as in JP}
	\end{align} 
	where 
	\begin{equation}
		\label{det7}
		{\xi_i^{SGD}=g(X_i,\theta_{i-1}^{LDP}) - R( \theta_{i-1}^{LDP})}
	\end{equation}
	captures the error due to measurement and  
	\begin{equation}
		\label{det8}
		{\xi_i^{LDP}= \mathcal{A}(g(X_i,\theta_{i-1}^{LDP}))-g(X_i,\theta_{i-1}^{LDP})}
	\end{equation}
	captures the {\it error due to privacy}.
	Analog to \eqref{det5}, the final ldp SDG estimate is defined by  
	\begin{align}
		\bar{\theta}^{LDP}&=\frac{1}{n}\sum_{i=1}^n\theta_i^{LDP}. \label{ldp sgd}
	\end{align}
	
	\begin{rem}
		\citet{song2013} also investigate mini-batch SGD under DP, where each iteration is updated as in \eqref{det6} and demonstrated that mini-batch SGD with batchsize $5$ achieves higher accuracy than batchsize $1$. 
		Their procedure then reads
		\begin{align*}
			\tilde{\theta}_i^{DP} =  \tilde{\theta}_{i-1}^{DP}-\eta_i \Big (\frac{1}{s}\sum_{j=1}^sg(X_{i_j},\tilde{\theta}_{i-1}^{DP})+ \frac{L_i}{s}\Big) =  \tilde{\theta}_{i-1}^{DP}-\eta_i \big(R(\tilde{\theta}_{i-1}^{(DP)}) + \xi_i^{SGD} + \xi_i^{DP}\big),
		\end{align*}
		where $L_i\sim \Lap(0,\Delta(g)/\epsilon)$, $\xi_i^{SGD}=\frac{1}{s}\sum_{j=1}^s g(X_{i_j},\tilde{\theta}_{i-1}^{DP}) - R(\tilde{\theta}_{i-1}^{DP})$ and $\xi_i^{DP}=L_i/s$. 
		Therefore, all results presented in this paper for the LDP-SGD also hold for the DP mini-batch SGD proposed by \citet{song2013}.
		However, this mini-batch SGD guarantees DP, not LDP. 
		To obtain an ldp mini-batch SGD, one would need to privatize each data, leading to the following iteration:
		\begin{align*}
			\tilde{\theta}_i^{LDP}=&\tilde{\theta}_{i-1}^{LDP}-\eta_i \frac{1}{s}\sum_{j=1}^s\big (g(X_{i_j},\tilde{\theta}_{i-1}^{LDP})+ L_{i_j}\big), 
		\end{align*}
		where $L_{i_1},\ldots,L_{i_s}\sim \Lap(0,\Delta(g)\epsilon)$ are independent random variables.
		Our results are applicable for  ldp mini batch SGD as well.
	\end{rem}
	\begin{rem}
		Note that the deconvolution approach  used  in 
		\citet{wang2022} for the construction of a DP bootstrap 
		procedure is not applicable for SGD, because it requires an additive relation  of the form $U=V+W$ between a DP-private and a non-private estimator $U$ and $V$, where the distribution  of $W$ is known.
		For example, if the gradient is linear, that is $R(\theta)= G\theta$ for a matrix $G$,  and SGD and LDP-SDG are started with the same initial 
		value $\theta_0$, we  obtain  by standard calculations the representation 
		\begin{align*}
			\bar{\theta}^{LDP} &= \bar{\theta}  + Z_n~~~
		\end{align*}
		where  $Z_n= -\frac{1}{n}\sum_{i=1}^n\sum_{j=1}^i \eta_j\xi_j^{LDP}\prod_{k=j+1}^i(1-G\eta_k) $. However,
		although the distribution of the random variables $\xi_j^{LDP}$ is known, the distribution of $Z_n$ is not easily accessible and additionally depends on the unknown matrix $G$, which makes the application of deconvolution principles not possible.
	\end{rem}

	\begin{algorithm}[H]
		\centering
		\caption{Local Differentially Private Stochastic Gradient Descent (LDP-SGD)}
		\begin{algorithmic}
			\State \textbf{Input: } $X_1,\ldots, X_n$, noisy gradient $g(\cdot, \cdot)$, $\theta_0$, learning rate 
			$\eta_i=ci^{-\gamma}$  with  parameters $c>0$ and $\gamma \in (0.5,1) $
			, $\epsilon$-ldp mechanism $\mathcal{A}$
			\State \textbf{Output: } $\theta_1^{LDP},\ldots,\theta_n^{LDP}$, $\bar{\theta}^{LDP}$
			\State $\bar{\theta}^{LDP}\gets 0$
			\For{$i=1,\ldots,n$}
			\State $\theta_i^{LDP} \gets \theta_{i-1}^{LDP}+ci^{-\gamma}\mathcal{A}(g(X_i,\theta_{i-1}^{LDP}))$
			\State $\bar{\theta}^{LDP} \gets \frac{i-1}{i}\bar{\theta}^{LDP} + \frac{1}{i}\theta_i^{LDP}$
			\EndFor
		\end{algorithmic}
	\end{algorithm}
	
	\noindent
	{\bf  Block Bootstrap.}
	Bootstrap is a widely used procedure to estimate   a distribution of an estimator
	$\hat \theta=\theta(X_1,\ldots,X_n)$   calculated from data $X_1,\ldots,X_n$  \cite[see][]{efron1994}.
	In the simplest case $X_1^\star, \ldots , X_n^\star$ are drawn  with replacement from   $X_1, \ldots , X_n$  and used to calculate  
	$\hat{ \theta}^\star =\theta(X_1^\star,\ldots,X_n^\star)$. 
	This procedure is repeated  several  times to obtain  an approximation  of the distribution $\hat{\theta}$. While it  provides a valid approximation in many (but not in all) cases if  $X_1, \ldots , X_n$ are independent identically distributed, the bootstrap approximation is not correct in the case of dependent data.
	A common approach to address this problem is the multiplier bootstrap which is tailored to estimators with a linear structure \citep{lahiri2003}. 
	For illustration  of the principle, we consider the empirical mean $\hat {\theta} = \bar{X}_n =  \frac{1}{n} \sum_{i=1}^n X_i$. 
	Under suitable assumptions the central limit theorem for stationary sequences shows that for large sample size $n$ the distribution of $\sqrt{n} (\bar X_n -  \mu )  $ can be approximated by a normal distribution, 
	say ${\cal N} (0, \sigma^2) $,  
	with expectation $0$ and variance $\sigma^2$, 
	where $\mu = \E[X_1]$ and $\sigma^2 $ depends in a complicated manner on the dependence structure of  the data.  \\
	The multiplier block bootstrap mimics this distribution by partitioning the sample into $m$ blocks of length $l$, say $\{X_{(i-1)l+1},\ldots,X_{il}\}$.
	For each block one calculates the mean $\bar{X}_{(i)}=\frac{1}{l}\sum_{j=(i-1)l+1}^{il} X_j$ which is then multiplied with a random weight $\epsilon_i$ with mean $0$ and variance $1$ to obtain the estimate $\bar{X}^\star_n = \frac{1}{m}\sum_{i=1}^m \epsilon_i \bar{X}_{(i)}$ where $m=\lfloor \frac{n}{l}\rfloor$. 
	If the dependencies between $X_i$ and $X_j$ become sufficiently small  for increasing $| i- j | $, it can be shown that  the distribution of $\sqrt{n} (\bar{X}_n^\star -  \bar{X}_n )  $ is a reasonable approximation  of the distribution $\sqrt{n} (\bar X_n -  \mu ) $ (for large $n$).
	Typical conditions on the dependence structure of the data guaranteeing these approximations are formulated in terms of mixing or physical dependence coefficients \cite[see][]{bradley2005,Wu2005}.
	We will not give details here, because none of these techniques will be applicable for proving the validity of the multiplier block bootstrap developed in the following section.

	\section{Multiplier Block Bootstrap for LDP-SGD} \label{main}
	
	In this section, we will develop  a multiplier block bootstrap approach to obtain 
	the distribution of the LDP-SGD  estimate  $ \bar{\theta}^{LDP}$  defined  
	in  \eqref{ldp sgd} by resampling. 
	We also prove that this method is statistically valid in the sense that 
	the bootstrap distribution converges weakly (conditional on the data) to the  limit distribution of the estimate $ \bar{\theta}^{LDP}$, which is derived first.
	
	Our first result is a direct consequence  of Theorem 2 in  \citet{Polyak1992} which can be applied to LDP-SGD.
	
	\begin{thm} \label{DP SGD normal nonlinear}
		If \cref{conditions Polyak} 
		in the supplement holds, then the LDP-SGD estimate   $\bar{\theta}^{LDP}$  in \eqref{ldp sgd}  satisfies
		\begin{align}
			\label{det10}
			\sqrt{n}(\bar{\theta}^{LDP}-\theta_\star)\overset{d}{\rightarrow} N(0,\Sigma),
		\end{align}
		where the matrix $\Sigma$ is given by
		$\Sigma = G^{-1} S G^{-1}$. Here $S$ is the asymptotic variance of the errors $\xi_i=\xi_i^{SGD}+\xi_i^{LDP}$ and $G$ is a linear approximation of $R(\theta)$ (see \cref{conditions Polyak} in the supplement for more details).
	\end{thm}
	
	\begin{rem} ~~\\
		(a) \cref{conditions Polyak} requires the sequence of  errors $\{\xi_i\}_{i\geq 1}$ to be a martingale difference process 
		with respect to a filtration $\{\mathcal{F}_i\}_{i\geq 0}$. This assumption is satisfied,  if $\{\xi_i^{SGD}\}_{i\geq 1}$  and $\{\xi^{LDP}_i\}_{i\geq 1}$ are 
		martingale difference processes  with respect to $\{\mathcal{F}_i\}_{i\geq 0}$.   
		Note that the condition 
		$\E[\xi_i^{LDP}|\mathcal{F}_{i-1}]=0$ is implied by 
		$\E[\mathcal{A}(g(X,\theta))|X]=g(X,\theta)$, which is satisfied for the Laplace mechanism and Randomized Response can be adjusted to fulfill this requirement.\\ 
		(b) 
		If $\xi_t^{LDP}$ and $\xi_t^{SGD}$ are independent given $\mathcal{F}_{t-1}$ and their respective covariance matrices converge in probability to $S_{LDP}$ and $S_{SGD}$, then it holds that $S=S_{SGD} + S_{LDP}$.
	\end{rem}

	In principle ldp statistical inference 
	based on  the  statistic $\bar{\theta}^{LDP}$ can be made using the asymptotic distribution in Theorem \ref{DP SGD normal nonlinear} with an ldp estimator of the covariance  matrix 
	$\Sigma$ in \eqref{det10}. 
	For this purpose the  matrices  $G$ and $S$ have to be estimated in an ldp way.
	While $S$ can be estimated directly from the ldp observations $\mathcal{A}(g(X_i,\theta_{i-1}^{LDP}))$ in \eqref{DP as in JP}  
	by
	\begin{equation*}
		\bar{S}^{LDP}=\frac{1}{n}\sum_{i=1}^n\mathcal{A}(g(X_i,\theta_{i-1}^{LDP}))(\mathcal{A}(g(X_i,\theta_{i-1}^{LDP})))^T,
	\end{equation*}
	the matrix $G$ 
	has to be estimated separately. Therefore, the privacy budget has to be split on the estimation of $G$ and on the iterations of SGD.
	As a consequence this approach 
	leads to high inaccuracies since all components need to be estimated in a ldp way.
	Furthermore, common estimates of $G$ are based on  the derivative of $g$ (with respect to $\theta$), and 
	therefore $g$ needs to be differentiable. This excludes important examples such as quantile regression.
	Additionally, the computation of the inverse of $G$ can become computationally costly in high dimensions.
	
	As an alternative we will develop a multiplier block bootstrap, which avoids these problems. 
	Before we explain our approach in detail we note that the bootstrap method proposed in 
	\citet{fang2018} for  non private SGD  is not applicable in the present context.
	These authors replace the recursion in \eqref{det4} by 
	$\theta_i^{\star}=\theta_{i-1}^\star - \eta_i w_i g(X_i,\theta_{i-1}^\star)$ where $w_1 , \ldots , w_n$ are independent identically distributed non-negative random variables with $Var(w_i)=1=\E[w_i]$. By applying SGD in this way $B$ times they calculate 
	SGD estimates $\bar{\theta}^{\star(1)} , \ldots , \bar{\theta}^{\star(B)}$, which are used to estimate the distribution of $\bar \theta $. However, for 
	the implementation of this approach under LDP
	the privacy budget $\epsilon$ must be split onto these $B$ versions.
	In other words, for the calculation of 
	each bootstrap  replication  $\bar{\theta}^{\star(b)}$ there is only a privacy budget of $\epsilon/B$  left, resulting in highly inaccurate estimators.
	\\
	To address these problems we propose to apply the multiplier block bootstrap principle to the iterations $\theta_1^{LDP},\ldots,\theta_n^{LDP}$ of LDP-SGD in order to mimic the strong dependencies in this data. 
	To be precise let $l=l(n)$ be the block length and
	$m=\lfloor\frac{n}{l}\rfloor$ the number of blocks.
	Let $\epsilon_1,\ldots,\epsilon_m$ be bounded, real-valued, independent and identically distributed  random variables with $\E[\epsilon_i]=0$ and  equal to ${\rm Var} (\epsilon_i)= 1$. 
	For the given iterates $\theta_1^{LDP},\ldots,\theta_n^{LDP}$ by LDP-SGD in \eqref{DP as in JP} we calculate a bootstrap analog of $\bar{\theta}^{LPD}-\theta_\star$ as 
	\begin{equation} \label{bootstraped estimator}
		\bar{\theta}^{\star}=
		\frac{1}{ml}\sum_{j=1}^m\epsilon_j\sum_{b=(j-1)l+1}^{jl}(\theta_b^{LDP}-\bar{\theta}^{LDP}),
	\end{equation}
	where  $\bar{\theta}^{LDP}$ is the LDP-SGD estimate defined in \eqref{ldp sgd}.
	We will show below that under appropriate assumptions the distribution of $\bar{\theta}^\star$ is a reasonable approximation of the distribution of $\bar{\theta}^{LDP}-\theta_\star$. 
	In practice this distribution can be obtained by generating $B$ samples of the form \eqref{bootstraped estimator}.
	Details are given in \cref{algo block bootstrap}, where the multiplier block bootstrap is used to construct an $\alpha$-quantile for the distribution of $\bar{\theta}^{LDP}- \theta_\star$, say $q_{\alpha}$. With these quantiles we obtain an $(1-2\alpha)$ confidence interval for $\theta_\star$ as
	\begin{equation}\label{CI}
		\hat{C}=[\bar{\theta}^{LDP}+q_{\alpha},\bar{\theta}^{LDP}+q_{1-\alpha}].
	\end{equation}

	\begin{algorithm}
		\centering
		\caption{Block Bootstrap for Stochastic Gradient Descent}
		\label{algo block bootstrap}
		\begin{algorithmic}
			\State \textbf{Input: } $\theta_1,\ldots,\theta_n$, $B$, $l$, $\alpha  \in (0,1)$ 
			\State \textbf{Output: } an $\alpha $-quantile of the distribution of $\bar{\theta}^{LDP}- \theta_\star$ 
			\State Set $\bar{\theta} = \frac{1}{n} \sum_{i=1}^n\theta_i$
			and $m = \lfloor\frac{n}{l}\rfloor$
			\For{$i=1...B$}
			\State Draw $\epsilon_1,\ldots,\epsilon_m$ independent and identically distributed with $\E[\epsilon_i]=0$ and $Var(\epsilon_i)=1$
			\State $\bar{\theta}^{\star(i)} \gets \frac{1}{ml}\sum_{j=1}^m\epsilon_j\sum_{b=(j-1)l+1}^{jl}(\theta_b-\bar{\theta})$
			\EndFor
			\State $q_{\alpha}\gets$ empirical $\alpha$ quantile of $\{\bar{\theta}^{\star(1)},\ldots,\bar{\theta}^{\star(B)}\}$
		\end{algorithmic}
	\end{algorithm}

	\begin{thm}
		\label{BB is consistent nonlinear}
		Let \cref{conditions Polyak,conditions LR} in the supplement be satisfied. 
		Let $l=l(n)$ and $m=m(n)$ such that $l, m\to \infty$ and $m^\gamma l^{\gamma-1}\to 0$ for $n\to \infty$. 
		Let $\epsilon_1,\ldots,\epsilon_m$ in \eqref{bootstraped estimator} be independent identical distributed random variables independent of $\theta_1^{LDP},\ldots, \theta_n^{LDP}$ with $\E[\epsilon_1]=0$ and $Var(\epsilon_1)=1$. 
		Further assume that there exists a constant $C$ such that $|\epsilon_i|\leq C$ a.s.. 
		Then, conditionally on $\theta_1^{LDP},\ldots,\theta_n^{LDP}$,
		\[\sqrt{n}\bar{\theta}^{\star}\overset{d}{\rightarrow} N(0,\Sigma)\]
		in probability with $\Sigma = G^{-1}SG^{-1}$.
		Here $S$ is the asymptotic variance of the errors $\xi_i=\xi_i^{SGD}+\xi_i^{LDP}$ and $G$ is a linear approximation of $R(\theta)$ (see \cref{conditions Polyak} in the supplement for more details).
	\end{thm}
	\begin{rem} ~~\\
		(a)
		If we chose $l=\lfloor n^\beta \rfloor$ (which yields $m=\lfloor n^{1-\beta}\rfloor$), $l$ and $m$ satisfy the assumptions of \cref{BB is consistent nonlinear} if $\gamma<\beta$.
		The parameter $\beta$ determines the number of blocks used in the multiplier bootstrap. 
		On the one hand we would like to have as many blocks as possible since we expect better results if more samples are available. 
		This would suggest to choose $\beta$ close to $\gamma$. 
		On the other hand, $\beta$ also determines the block-length $l$, which needs to be large enough to capture the underlying strong dependence structure of the iterates of LDP-SGD. 
		This suggests to choose $\beta$ close to one.\\
		(b)  If the blocks have been calculated, the run time of the block bootstrap is of order $O(Bmd)$, which is dominated by the run time $O(nd)$ of (LDP-)SGD, as long as $B=o(l)$.
	\end{rem}

	\section{Simulation}
	We will consider LDP-SGD for the estimation of a quantile and  of the parameters in a quantile regression model.
	We investigate these models because here the gradient is not differentiable and the plug-in estimator from \citet{chen2020} can not be used.
	In each scenario, we consider samples of size $n=10^6, 10^7, 10^8$ and privacy parameter $\epsilon=1$. 
	The stepsize of the SGD is chosen as $\eta_i=i^{-0.51}$. 
	The $90\%$ confidence intervals \eqref{CI} for the quantities  of interest are computed by  $B=500$ bootstrap replications. 
	The block length is chosen as $l=\lfloor n^\beta\rfloor$, where $\beta=0.75$. 
	The distribution of multiplier is chosen as $\epsilon_i\sim \Unif(-\sqrt{3},\sqrt{3})$. 
	The empirical coverage probability and average length of the interval $\hat{C}$ are estimated by $500$ simulation runs.
	All simulations are run in R \citep{R} and executed on an internal cluster with an AMD EPYC 7763 64-Core processor under Ubuntu 22.04.4.
	
\begin{figure}[h]
		\centering
		\includegraphics[width=0.45\textwidth]{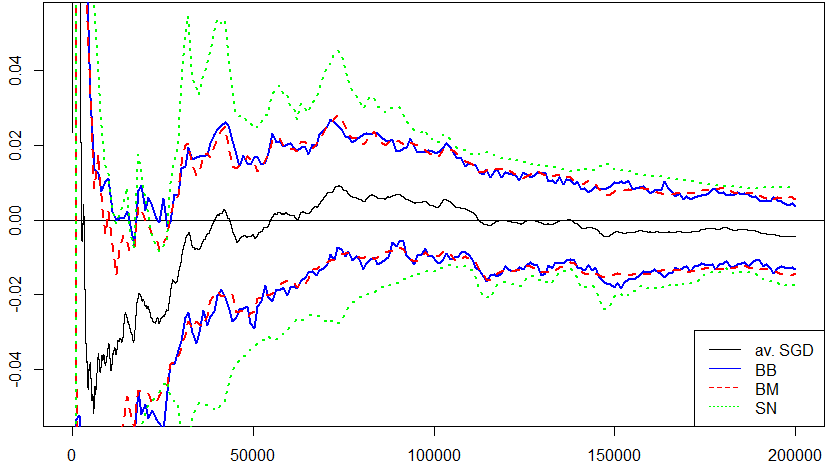} \hspace{0.05\textwidth}\includegraphics[width=0.45\textwidth]{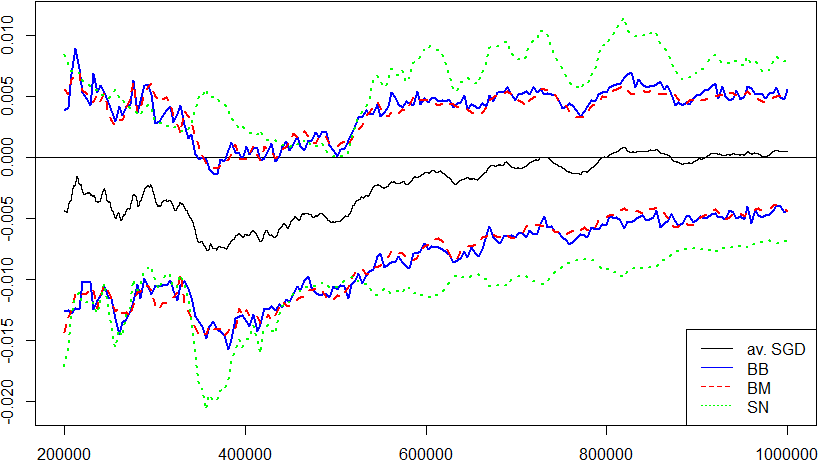}
		
		\caption{Trajectory of the LDP-SGD estimate $\bar{\theta}_n^{LDP}$ (av. SGD) and the upper and lower boundaries of confidence intervals obtained by block bootstrap (BB), batch mean (BM) and self normalisation (SN) 
			of the $50\%$ quantile of a standard normal distribution. The left and right panel correspond to different sample sizes, left: $n=10 - 200 000$; right: $n=200 000 - 1 000 000$.}
		
		\label{Trajectory CI}
	\end{figure}

	{\bf Quantiles:}
	Let $X_1,\ldots,X_n$ be independent and identically distributed $\mathbb{R}$ valued random variables with distribution function $F_X$ and density $p_X$. Denote by $x_\tau$ the $\tau$-th quantile of $X_1$ i.e. $F_X(x_\tau)=\tau$. We assume that $p_X(x)>0$ in a neighborhood of $x_\tau$, then $x_\tau$ is the (unique) root of the equation
	$
	R(\theta)
	= -\tau +F_X(\theta).
	$
	The noisy gradient is given by
	$g(X_i,\theta)=
	-\tau+\mathbb{I}\{X_i\leq \theta\},$
	which can be privatized using Randomized Response, that is 
	\[\mathcal{A}(g(X_i,\theta))=-\tau + \frac{\mathcal{A}_{RR}(\mathbb{I}\{X_i\leq \theta\})}{2p-1}-\frac{1-p}{2p-1},\]
	where $\mathcal{A}_{RR}$ is defined in Example \ref{Randomized Response} and $p=\frac{e^\epsilon}{1+e^\epsilon}$. Note that the re-scaling ensures that $\E[\mathcal{A}(g(X_i,\theta))|X_i]=g(X_i,\theta)$ for all $\theta \in \mathbb{R}$.
	The measurement error and the error due to privacy in \eqref{det7} and \eqref{det8} are given by
	\begin{align*}
		\xi_i^{SGD}& =\mathbb{I}\{X_i\leq \theta_{i-1}^{LDP}\}-F_X(\theta_{i-1}^{LDP})\text{ , }\\
		\xi_i^{LDP} &= \frac{\mathcal{A}_{RR}(\mathbb{I}\{X_i\leq \theta_{i-1}^{LDP}\})}{2p-1}-\frac{1-p}{2p-1} - \mathbb{I}\{X_i\leq \theta_{i-1}^{LDP}\},
	\end{align*}
	respectively, which define indeed martingale difference processes with respect to the filtration $\{\mathcal{F}_i\}_{i\geq1}$, where $\mathcal{F}_i = \sigma(\theta_{1}^{LDP},\ldots, \theta_{i}^{LDP})$ and $\sigma(Y)$ denotes the sigma algebra generated by the random variable $Y$.
	The assumptions of \cref{BB is consistent nonlinear,DP SGD normal nonlinear} can be verified 
	with $\Sigma=(S_{SGD}+S_{LDP})/(p_X(x_\tau))^2$ where
	\begin{align*}
		S_{SGD}&=F_X(x_\tau)(1-F_X(x_\tau))=\tau(1-\tau)\text{, } 
		& S_{LDP}=\tfrac{e^\epsilon}{e^\epsilon-1}.
	\end{align*}
	In \cref{simulation quantile} we display the simulated coverage probabilities and length of the confidence intervals for the $50\%$ and $90\%$ quantile of a standard normal distribution calculated by block bootstrap (BB). 
	We also compare our approach with the batch mean (BM) introduced in \citet{chen2020} and the self normalisation (SN) approach suggested in \citet{liu2023online}. 
	We observe that BB and BM behave very similar with respect to the empirical coverage and the length of the computed confidence intervals while the confidence intervals obtained by SN are slightly larger.
	\\
	In \cref{Trajectory CI} we display the trajectory of the upper and lower boundaries of a $90\%$ confidence interval for the $50\%$ quantile of a standard normal distribution for the BB, BM and SN approach.
	Again, we observe that the confidence intervals obtained by BB and BM are quite similar and the confidence intervals obtained by SN are wider.
	
	\begin{table}[h]
		\centering 
		
		\caption{Empirical coverage and length of a $90\%$ confidence interval for the $50\%$ and $90\%$ quantile of a standard normal distribution. The confidence intervals are obtained by LDP-SGD with self normalization (SN), batch mean (BM) and the block bootstrap (BB) proposed in this paper.
			The numbers in brackets are the standard errors. }
		
		\label{simulation quantile}
		\begin{tabular}{c cc c c c }
			$\tau$ & & & $10^6$ & $10^7$ & $10^8$  \\
			\hline
			\multirow{6}{*}{0.5} & \multirow{2}{*}{BB}   & cover: & 0.880 (0.015)  &  0.886 (0.014) &   0.886 (0.014)\\
			& & length: & 0.0085 $(5.2\times 10^{-5})$ & 0.0027 $(1.2 \times 10^{-5})$ & 0.0009 $(3.1 \times 10^{-6})$ \\ 
			\cline{2-6}
			& \multirow{2}{*}{BM}   & cover: & 0.884 (0.014) & 0.898 (0.014) & 0.896 (0.014) \\
			& & length: & 0.0086 $(5.2\times 10^{-5})$ & 0.0027 ($1.1\times 10^{-5})$ &  0.0009 $(2.9\times 10^{-6})$ \\   
			\cline{2-6}
			& \multirow{2}{*}{SN}   & cover: & 0.914 (0.013) &  0.898 (0.014) & 0.886 (0.014) \\
			& & length: & 0.0106 $(1.9\times 10^{-4})$ & 0.0034 $(6 \times 10^{-5})$ & 0.0011 $(2.1\times 10^{-5})$\\  
			
			\hline
			\hline
			\multirow{6}{*}{0.9} & \multirow{2}{*}{BB}   & cover: & 0.828 (0.017) &  0.856 (0.016) & 0.848 (0.016) \\
			& & length: & 0.0175 ($1.1\times 10^{-4}$) &0.0057 ($2.7\times 10^{-5}$) & 0.0018 ($6.4\times 10^{-6}$)\\ 
			\cline{2-6}
			& \multirow{2}{*}{BM}   & cover: &  0.830 (0.017) & 0.868 (0.015) & 0.840 (0.016) \\
			& & length: & 0.0179 ($1.1\times 10^{-4}$) & 0.0058 ($2.5\times 10^{-5}$) &  0.0018 ($5.8\times 10^{-6}$)\\   
			\cline{2-6}
			& \multirow{2}{*}{SN}   & cover: & 0.862 (0.015) &0.880 (0.015) &0.878 (0.015) \\
			& & length: & 0.0235 ($4.5\times 10^{-4}$) &0.0076 ($1.4\times 10^{-4}$) & 0.0024 ($4.3\times 10^{-5}$)\\  
		\end{tabular}
		
	\end{table}
	
	{\bf Quantile Regression:} Let $Z_1,\ldots,Z_n$ be iid random vectors where 
	$Z_i=(X_i,y_i)$, $X_i\in\mathbb{R}^d$ with $\Sigma_X= \E[X_iX_i^T]$ and 
	$y_i=X_i^T\beta_\tau + \varepsilon_i$ 
	where $\varepsilon_i$ is independent of $X_i$  with distribution function $F_\varepsilon$. 
	We further assume that $F_\varepsilon$ has a density in a neighbourhood of $0$, say $p_\varepsilon$, with  $p_\varepsilon(0)>0$, that $F_\varepsilon(0)=\tau$ and that there exists a constant $m>0$ such that $|X_i|\leq m$ for all $i=1,\ldots,d$. 
	If $Q_{y|X}(\tau)$ is the conditional quantile of $y$ given $X$, it follows that $Q_{y|X}(\tau)=X^T\beta_\tau$ and $R(\beta_\tau)=0$, where
	\[R(\beta)=-\tau \E[X] + \E[\mathbb{I}\{y-X^T\beta < 0\}X]\]
	which can be linearly approximated by $G=\Sigma_X p_\varepsilon(0)$, 
	since a Taylor expansion at $\beta_\tau$ yields
	\[\E[\mathbb{I}\{y-X^T\beta < 0\}|X]=\tau + p_\varepsilon(0)X^T(\beta-\beta_\tau) + O(\lVert\beta-\beta_\tau\rVert^2).\]
	The noisy observations are given by
	$g(Z_j,\beta)=(-\tau + \mathbb{I}\{y_j - X_j^T\beta\leq 0\})X_j$
	and can be privatized by the Laplace mechanism (see Example \ref{Laplace})
	\[\mathcal{A}_{Lap}(g(Z_j,\beta))= g(Z_j,\beta) + \mathbf{L}_j,\]
	where $\mathbf{L}_j=(L_1,\ldots,L_d)^T$ is a vector of independent and identical distributed Laplacian random variables, i.e. $L_i \sim \Lap(0,2\max(\tau,1-\tau)md/\epsilon)$.\\
	The measurement error and the error due to privacy in \eqref{det7} and \eqref{det8} are given by
	\begin{align*}
		\xi_i^{SGD}&=-\tau(X_i - \E[X]) + \mathbb{I}\{y_i-X_i\beta_{i-1}^{LDP} < 0\}X_i - \E[\mathbb{I}\{y-X\beta_{i-1}^{LDP} < 0\}X]\\
		\xi_i^{LDP} &= (L_1,\ldots,L_d)^T,
	\end{align*}
	which are indeed martingale difference processes with respect to the  filtration $\{\mathcal{F}_i\}_{i\geq1}$, where $\mathcal{F}_i = \sigma(\beta_{1}^{LDP},\ldots, \beta_{i}^{LDP})$.
	The assumptions from \cref{BB is consistent nonlinear,DP SGD normal nonlinear} can be verified 
	with $\Sigma=\Sigma_X^{-1} (S_{SGD}+S_{LDP}) \Sigma_X^{-1}/p_\varepsilon(0)^2$ and
	\begin{align*}
		S_{SGD}&=\tau(1-\tau)\Sigma_X \text{, }
		& S_{LDP}=2 \tfrac{4\max\{\tau^2,(1-\tau)^2\}m^2d^2}{\epsilon^2}\mathbb{I}_{d\times d}.
	\end{align*}
	
	We compare the confidence intervals obtained by BB and BM where $X=(1,X_1,X_2,X_3)$, $(X_1,X_2,X_3)$ follows a truncated standard normal distribution on the cube $[-1,1]^3$ and $\varepsilon_i$ is normal distributed with variance $1$.
	For the generation of the truncated normal and the Laplace distribution we used the package of \cite{TruncatedNormal} and \cite{ExtDist}, respectively.
	The simulation results are displayed in \cref{simulation QR}. 
	We observe that for $n=10^6$ BB yields too small and BM yields too large coverage probabilities, while the lengths of the confidence intervals from BB are smaller. 
	If $n=10^7$ and $n=10^8$ the coverage probabilities from BB and BM are increasing and decreasing respectively.
	
	\begin{table}[h]
		\centering
		
		\caption{Empirical coverage and length of a $90\%$ confidence interval for the parameters of quantile regression with $\beta_\tau=(0,0,1,-1)$ where $\tau=0.5$. 
			The confidence intervals are obtained by batch mean (BM) and the block bootstrap (BB) proposed in this paper.
			The numbers in brackets are the standard errors. }
		
		\label{simulation QR}
		\begin{tabular}{c cc c c c}
			& & & $10^6$ & $10^7$ & $10^8$  \\
			\hline
			\multirow{8}{*}{BB} & \multirow{2}{*}{$\beta_0$} & cover: & 0.860 (0.016)  &  0.882 (0.014)  &  0.904 (0.013)\\
			& & length: & 0.07 $(1.2 \times 10^{-3})$  & 0.016 $(1.2 \times 10^{-4})$  & 0.005 $(1.7 \times 10^{-5})$\\ 
			\cdashline{2-6}
			& \multirow{2}{*}{$\beta_1$} & cover: & 0.862 (0.015) & 0.868 (0.015)  &  0.880 (0.015) \\
			& & length: & 0.228 $(6.3 \times 10^{-3})$& 0.055 $(4.8 \times 10^{-4})$& 0.016 $(6.2 \times 10^{-5})$\\ 
			\cdashline{2-6}
			& \multirow{2}{*}{$\beta_2$} & cover: & 0.850 (0.016) & 0.870 (0.015) & 0.882 (0.014)\\
			& & length: & 0.241 $(5.6 \times 10^{-3})$& 0.054 $(5.3 \times 10^{-4})$& 0.016 $(6.9 \times 10^{-5})$\\
			\cdashline{2-6}
			& \multirow{2}{*}{$\beta_3$} & cover: & 0.844 (0.016) & 0.866 (0.015) & 0.890 (0.014) \\
			& & length: & 0.243 $(5.6 \times 10^{-3})$& 0.054 $(8.0 \times 10^{-4})$ & 0.016$(6.7 \times 10^{-5})$\\ 
			\hline
			\multirow{8}{*}{BM} & \multirow{2}{*}{$\beta_0$} & cover: & 0.952 (0.01)  &  0.932 (0.011)  &   0.918 (0.012)\\
			& & length: & 0.103 $(2.4 \times 10^{-3})$ & 0.02 $(3.6 \times 10^{-4})$ & 0.005$(2.1 \times 10^{-5})$\\ 
			\cdashline{2-6}
			& \multirow{2}{*}{$\beta_1$} & cover: & 0.938 (0.012) & 0.928 (0.012) &  0.896 (0.014) \\
			& & length: & 0.315 $(9.4 \times 10^{-3})$ & 0.076 $(1.5 \times 10^{-3})$& 0.017 $(1.3 \times 10^{-4})$\\ 
			\cdashline{2-6}
			& \multirow{2}{*}{$\beta_2$} & cover: & 0.948 (0.01) & 0.926 (0.012) & 0.902 (0.013)\\
			& & length: & 0.335 $(8.4 \times 10^{-3})$& 0.073 $(1.7 \times 10^{-3})$ & 0.017 $(2.0 \times 10^{-4})$ \\
			\cdashline{2-6}
			& \multirow{2}{*}{$\beta_3$} & cover: & 0.928 (0.012) &0.928 (0.012) & 0.898 (0.014) \\
			& & length: & 0.339 $(8.5 \times 10^{-3})$ & 0.073 $(2.3 \times 10^{-3})$& 0.017 $(1.5 \times 10^{-4})$
		\end{tabular}
		
	\end{table}
	
	\newpage
	\acksection
	Funded by the Deutsche Forschungsgemeinschaft (DFG, German Research Foundation) under Germany's Excellence Strategy - EXC 2092 CASA - 390781972.
	\bibliography{literatur}
	
	\newpage
	\appendix
	
	\section{Technical assumptions
	}
	
	The theoretical results of this paper, in particular the consistency of the multiplier block bootstrap,
	are proved under the following assumptions. 
	
	\begin{cond} \label{conditions Polyak}
		Denote $\xi_i=\xi_i^{LDP} + \xi_i^{SGD}$. Assume that the following conditions hold:
		\begin{enumerate}
			\item There exists a differentiable function $V: \mathbb{R}^d\to \mathbb{R}$ with  gradient $\nabla V$  such that for some constants $\lambda>0$, $c'>0$, $L'>0$, $\varepsilon>0$ 
			\begin{align*}
				V(\theta) & \geq c'\lVert \theta \rVert ^2 ~~ \text{ for all }   \theta \in \mathbb{R}^d   \\
				\lVert \nabla V(\theta)-\nabla V(\theta') \rVert & \leq L'\lVert\theta-\theta'\rVert  ~~ \text{ for all } 
				\theta,\theta'\in \mathbb{R}^d   \\
				\nabla V(\theta-\theta_\star)^TR(\theta) & >0  ~~ \text{ for all }  \theta\neq \theta_\star \\ 
				\nabla V(\theta-\theta_\star)^TR(\theta) & \geq \lambda V(\theta-\theta_\star) 
				~~ \text{ for all }   \theta \in \mathbb{R}^d   \text{ with } \lVert\theta-\theta_\star\rVert\leq \varepsilon
			\end{align*} 
			\item There exist positive definite matrix $G\in \mathbb{R}^{d\times d}$ and constants $K_1<\infty$, $\varepsilon>0$, $0<\lambda\leq 1$ such that for all $\lVert \theta-\theta_\star\rVert\leq \varepsilon$
			\begin{equation} \label{this is G}
				\lVert R(\theta)-G(\theta-\theta_\star)\rVert \leq K_1 \lVert\theta-\theta_\star\rVert^{1+\lambda}.
			\end{equation}
			\item $\{\xi_i\}_{i\geq 1}$ is a martingale-difference process with respect to a filtration $\{\mathcal{F}_i\}_{i\geq 0}$, and for a constant $K_2 > 0 $ it holds that for all $i\geq 1$
			\begin{align*}
				\E\left[\lVert \xi_i \rVert ^2|\mathcal{F}_{i-1}\right] +\lVert R(\theta_{i-1}^{LDP})\rVert ^2 \leq K_2 (1+\lVert \theta_{i-1}^{LDP}\rVert^2).
			\end{align*}
			\item For the errors $\{\xi_i\}_{i \ge 1}$, there exists a  decomposition of the form 
			$\xi_i=\xi_i(0) + \zeta_i(\theta_{i-1}^{LDP}) ,  $
			a positive definite matrix $S$  and a function $\delta: \mathbb{R^d} \to \mathbb{R},$ which is continuous at the origin,  such that 
			\begin{align*}
				\E[\xi_i(0)|\mathcal{F}_{i-1}] & = 0  ~~ \text{  a.s.  }  \\
				\lim_{i\to\infty} \E[\xi_i(0)\xi_i(0)^T|\mathcal{F}_{i-1}]  & =  S  ~~\text{  in probability   } \\
				\lim_{C\to\infty}\sup_i \E[\lVert\xi_i(0)\rVert^2\mathbb{I}(\lVert\xi_i(0)\rVert >C)\rVert\mathcal{F}_{i-1}]   & =  0    ~~ \text{  in probability   } \\
				\E[\lVert \zeta_i(\theta_{i-1}^{LDP})\rVert ^2|\mathcal{F}_{i-1}] & \leq \delta (\theta_{i-1}^{LDP})
				~~ \text{  a.s.  }
			\end{align*} 
			\item The learning rates are given by $\eta_i=ci^{-\gamma}$ with $c>0$ and $0.5<\gamma<1$.
		\end{enumerate}
	\end{cond}

	\begin{cond}\label{conditions LR} ~~   
		Denote $\xi_i=\xi_i^{SGD}+\xi_i^{LDP}$ and assume that $\{\xi_i\}_{i\geq 1}$ is a martingale difference process with respect to a filtration $\{\mathcal{F}_i\}_{i\geq 0}$. Assume that the following holds:
		\begin{enumerate}
			\item There exists a function $s:\mathbb{R}^d\to\mathbb{R}^{d\times d}$ such that
			\[\E[\xi_i\xi_i^T|\mathcal{F}_{i-1}] = S + s(\theta_{i-1}^{LDP}-\theta_\star) ~~~\text{a.s.},\]
			where $S\in\mathbb{R}^{d\times d}$ is the matrix in \cref{conditions Polyak}, 
			and $\sqrt{tr(\E[s(\theta_i^{LDP}-\theta_\star)]^2)}\leq s_1 \sqrt{\eta_i}$ for a constant $s_1>0$.
			\item  There exists a constant $K<\infty$ such that $tr(\E[(\xi_i\xi_i^T)^2|\mathcal{F}_{i-1}])\leq K$ a.s..
			\item \ There exists a constant $\rho$ with $\frac{1}{m} \sum_{i=1}^m |\frac{1}{\sqrt{l}}\sum_{j=(i-1)l+1}^{il} \xi_j|^3\overset{\p}{\to} \rho$. 
			\item The sequence   $\{\xi_i \xi_i^T\}_{i \ge 1} $ is uniformly integrable.
			\item
			The learning rates are given by   $\eta_j=c j^{-\gamma}$ and 
			there exists a constant $b_0 \in \mathbb{N}$  such that $b_0^\gamma\geq 2\lambda(G)c$  and  
			\begin{equation} \label{base case}
				\sum_{j=1}^{b_0}\lambda(G)^2 \eta_j^2\prod_{k=j+1}^{b_0} (1-\lambda(G)\eta_k)^2\geq \lambda(G)\eta_{b_0}/2
			\end{equation}
			for every eigenvalue $\lambda(G)$ of the matrix $G$ in \eqref{this is G}.
		\end{enumerate}
	\end{cond}
	
	\begin{rem}
		In many cases, $tr(\E[s(\theta_i^{LDP}-\theta_\star)]^2)$ is approximately proportional to  $\E[\lVert\theta_i-\theta_\star\rVert_2]^2$ (which follows by a Taylor expansion) and the convergence rate of $\E[\lVert\theta_i-\theta_\star\rVert_2]$ is well studied, see for example \citet{chung1954stochastic} and \citet{chen2020}.
		If the sequence $\{\xi_i\}_{i\geq 1}$ is bounded, \cref{conditions LR} (2)-(4) are satisfied.\\
		Condition \eqref{base case} is the base case of a proof by induction and can be numerically verified for different choices of the parameters $c$ and $\gamma$ in the learning rate.
	\end{rem}
	
	\section{Proofs of the results in Section \ref{main}}

	We will denote $\xi_i:=\xi_i^{SGD} + \xi_i^{LDP}$ in this section. 
	Furthermore, we omit the superscript "LDP" of $\theta^{LDP}$ for improved readability. 
	In other words, we write $\theta_i$ for $\theta_i^{LDP}$ and $\bar{\theta}$ for $\bar{\theta}^{LDP}$. Moreover, for the sake of simplicity, we assume that $n=ml$.
	
	\subsection{Proof of Theorem \ref{DP SGD normal nonlinear}}
	This follows from  Theorem 2 in \citet{Polyak1992} by observing that  $\xi_i=\xi_i^{SGD}+\xi_i^{LDP}$ satisfies the assumptions stated in this reference.

	\subsection{Proof of Theorem \ref{BB is consistent nonlinear}}
	We first prove the statement in the case where the gradient $R(\theta)$ is assumed to be linear.
	After that, \cref{BB is consistent nonlinear} is derived by showing that bootstrap LDP-SGD in the non linear case is asymptotically approximated by a linearized version of bootstrap LDP-SGD. The following result is shown in Section \ref{proof BB is consisten linear}.

	\begin{thm} \label{BB is consistent linear}
		Assume that the conditions of \cref{BB is consistent nonlinear} hold and that $R(\theta)=G(\theta-\theta_\star)$ for a positive definite matrix $G\in\mathbb{R}^{d\times d}$.
		Then, conditionally on $\theta_1^{LDP},\ldots,\theta_n^{LDP}$
		\[\sqrt{n}\bar{\theta}^{\star}\overset{d}{\rightarrow} N(0,\Sigma)\]
		in probability, where $\Sigma = G^{-1}SG^{-1}$.
	\end{thm}
	From \cref{BB is consistent linear} we conclude that \cref{BB is consistent nonlinear} holds. 
	To be precise, let $G\in\mathbb{R}^{d\times d}$ be the matrix in the linear approximation of the gradient $R(\theta)$ in \eqref{this is G}, let $\{\xi_i\}_{i\geq1}$ be the martingale difference process capturing the error due to measurement and privacy as in \eqref{DP as in JP} and let $S\in\mathbb{R}^{d \times d}$ be the matrix in \cref{conditions Polyak} (4).
	Next we define a sequence of iterates $\theta_i^1$ of LDP-SGD for a loss function with $R(\theta)=G(\theta-\theta_\star)$, that is, $\theta_0^1=\theta_0$ and for $i\geq 1$
	\begin{align*}
		\theta_i^1 & = \theta_{i-1}^1 - \eta_i (G (\theta_{i-1}^1 -\theta_\star)+\xi_i) \\
		\bar{\theta}_n^1 & =\frac{1}{n}\sum_{i=1}^{n} \theta_i^1.\nonumber
	\end{align*}
	We denote the bootstrap analogue defined as in \eqref{bootstraped estimator} for this sequence by $\bar{\theta}^{1,\star}$, that is
	\begin{align*}
		\bar{\theta}^{1,\star} = \frac{1}{ml}\sum_{j=1}^m\epsilon_j\sum_{b=(j-1)l+1}^{jl}(\theta_b^{1}-\bar{\theta}^{1}_n).
	\end{align*}
	By \cref{BB is consistent linear} it follows conditionally on  $\theta_1^1,\ldots,\theta_n^1$,
	\[
	\sqrt{n}\bar{\theta}^{1,\star}\overset{d}{\rightarrow} N(0, G^{-1} S G^{-1})
	\]
	in probability.
	Since there is a bijection from $\theta_1^1,\ldots,\theta_n^1$ to $\xi_1,\ldots,\xi_n$ and from $\theta_1,\ldots,\theta_n$ to $\xi_1,\ldots,\xi_n$, the corresponding sigma algebras coincide and therefore, conditionally on $\theta_1,\ldots,\theta_n$,
	\[
	\sqrt{n}\bar{\theta}^{1,\star}\overset{d}{\rightarrow} N(0, G^{-1} S G^{-1})
	\]
	in probability.
	The assertion of \cref{BB is consistent nonlinear} is now a consequence of
	\[\sqrt{n}(\bar{\theta}^\star-\bar{\theta}^{1,\star})\overset{\p}{\rightarrow} 0.\] 
	For a proof of this statement, we define
	\begin{equation}
		k_j=i\mathbb{I}\{(i-1)l+1\leq j \leq il\} \label{def kj} 
	\end{equation}
	and note that
	\begin{align*}
		\sqrt{n}(\bar{\theta}^\star-\bar{\theta}^{1,\star}) &= 
		\sqrt{n} \sum_{i=1}^n\epsilon_{k_i}(\theta_i-\bar{\theta}-\theta_i^1+\bar{\theta}^1) \\
		&=\sqrt{n} \sum_{i=1}^n\epsilon_{k_i}(\theta_i-\theta_i^1)
		+ \sqrt{n}(\bar{\theta}^1-\theta_\star)\frac{1}{m}\sum_{i=1}^m\epsilon_i-\sqrt{n}(\bar{\theta}-\theta_\star) \frac{1}{m}\sum_{i=1}^m\epsilon_i.
	\end{align*}
	Here the second and third term converge to zero in probability, since $\frac{1}{m}\sum_{i=1}^m\epsilon_i$ does and both $\sqrt{n}(\bar{\theta}^1-\theta_\star)$, $\sqrt{n}(\bar{\theta}-\theta_\star)$ converge weakly by \cref{DP SGD normal nonlinear}. 
	For a corresponding statement regarding the first term let $I$ denote the $d$-dimensional identity matrix and note that
	\begin{align*}
		\theta_i-\theta_i^1 =& \theta_{i-1}-\eta_i R(\theta_{i-1}) -\eta_i \xi_i - (\theta_{i-1}^1 -\eta_iG(\theta_{i-1}^1 -\theta_\star) -\eta_i\xi_i)\\
		= & (I-\eta_i G) (\theta_{i-1}-\theta_{i-1}^1) - \eta_i ( R(\theta_{i-1}) - G(\theta_{i-1}-\theta_\star )) \\
		=& (I-\eta_i G)\big( (I-\eta_{i-1} G) (\theta_{i-2}-\theta_{i-2}^1) - \eta_{i-1} ( R(\theta_{i-2}) - G(\theta_{i-2}-\theta_\star ))\big)\\
		&- \eta_i ( R(\theta_{i-1}) - G(\theta_{i-1}-\theta_\star )) \\
		=& (I-\eta_i G)(I-\eta_{i-1} G) (\theta_{i-2}-\theta_{i-2}^1)\\
		&- (I-\eta_i G)\eta_{i-1} ( R(\theta_{i-2}) - G(\theta_{i-2}-\theta_\star )) - \eta_i ( R(\theta_{i-1}) - G(\theta_{i-1}-\theta_\star ))\\
		\vdots&\\
		=&  \Big(\prod_{k=1}^i(I-\eta_kG)\Big) (\theta_0 -\theta_0^1)- \sum_{j=1}^i \eta_j  \Big(\prod_{k=j+1}^i (I-\eta_kG)\Big) (R(\theta_{j-1})-G(\theta_{j-1}-\theta_\star))\\
		=& - \sum_{j=1}^i \eta_j \Big(\prod_{k=j+1}^i (I-\eta_kG)\Big) (R(\theta_{j-1})-G(\theta_{j-1}-\theta_\star)),
	\end{align*}
	since $\theta_0=\theta_0^1$.
	Therefore,  since $\max | \epsilon_j| \leq C$ a.s. by assumption, we obtain
	\begin{align*}
		\left\lVert\frac{1}{\sqrt{n}}\sum_{i=1}^n\epsilon_{k_i}(\theta_i-\theta_i^1) \right\rVert 
		=& \left\lVert \frac{1}{\sqrt{n}}\sum_{i=1}^n\epsilon_{k_i}\sum_{j=1}^i \eta_j \Big(\prod_{k=j+1}^i (I-\eta_kG)\Big) (R(\theta_{j-1})-G(\theta_{j-1}-\theta_\star))  \right\rVert \\
		\leq & \frac{C}{\sqrt{n}}\sum_{i=1}^n \left\lVert\sum_{j=1}^i \eta_j \Big(\prod_{k=j+1}^i (I-\eta_kG)\Big)  (R(\theta_{j-1})-G(\theta_{j-1}-\theta_\star)) \right\rVert\\
		=& \frac{C}{\sqrt{n}}\sum_{i=1}^n\eta_i \left\lVert\Big(\sum_{j=i}^n   \prod_{k=i+1}^j (I-\eta_kG)\Big)(R(\theta_{i-1})-G(\theta_{i-1}-\theta_\star))\right\rVert\\
		\leq &  \frac{C}{\sqrt{n}}\sum_{i=1}^n\eta_i\lVert R(\theta_{i-1})-G(\theta_{i-1}-\theta_\star)\rVert \left\lVert\sum_{j=i}^n   \prod_{k=i+1}^j (I-\eta_kG)\right\rVert_M,
	\end{align*}
	where $\lVert \cdot \rVert_M$ denotes a matrix norm.
	Following the same arguments as in \citet{Polyak1992}, part 4 of the proof of Theorem 2, it follows that the last term converges in probability to zero.

	\subsection{Proof of Theorem  \ref{BB is consistent linear}} \label{proof BB is consisten linear}
	The proof is a consequence of the following three Lemmas:
	\begin{lemma} \label{decomposition bootstrap}
		The bootstrap estimate $\bar{\theta}^\star$ in \eqref{bootstraped estimator} can be represented as 
		\begin{align} \label{eq decomposition bootstrap}
			\bar{\theta}^{\star}  =&\frac{1}{n\eta_1}\beta_1^n(\theta_0-\theta_\star)
			- \frac{1}{n}\sum_{j=1}^{n}\epsilon_{k_j}G^{-1}\xi_j 
			- \frac{1}{n}\sum_{j=1}^{n}\left(\beta_j^n -\epsilon_{k_j}G^{-1}\right)\xi_j 
			- (\bar{\theta}-\theta_\star)\frac{1}{m}\sum_{j=1}^{m}\epsilon_j,
		\end{align} 
		where the indices $k_j$ are defined in \eqref{def kj} and the $d\times d$ matrices $\beta_j^n$ are given by
		\begin{align}
			\beta_j^n&=\eta_j\sum_{i=j}^n\epsilon_{k_i}\prod_{k=j+1}^i(I-\eta_kG). \label{def betas}
		\end{align}
	\end{lemma}

	\begin{lemma} \label{convergence in distribution - new}
		Conditionally on 
		$\xi_1,\ldots,\xi_n$ we have
		\begin{equation*}
			\frac{1}{\sqrt{n}}\sum_{j=1}^{n}\epsilon_{k_j}G^{-1}\xi_j \overset{d}{\rightarrow} N(0,\Sigma)
		\end{equation*}
		in probability,
		where $\Sigma=G^{-1}S G^{-1}$.
	\end{lemma}

	\begin{lemma} \label{convergence in prop}
		Let the assumptions from \cref{BB is consistent linear} hold, $\beta_i^n$ as in \eqref{def betas} and $k_j$ as in \eqref{def kj}.
		Then
		\begin{equation*}
			\frac{1}{\sqrt{n}}\sum_{j=1}^{n}\left(\beta_j^n -\epsilon_{k_j}G^{-1}\right)\xi_j \overset{\p}{\rightarrow} 0.
		\end{equation*}
	\end{lemma}
	
	By the same arguments of Part 1 in the proof of Theorem 1 in \citet{Polyak1992} we obtain for the first term in \eqref{eq decomposition bootstrap} 
	$\frac{1}{\sqrt{n}\eta_1}(\theta_0-\theta_\star)\overset{\p}{\rightarrow} 0$ 
	(note that by assumption  $\epsilon_i$ are bounded a.s.).
	The fourth term is of order $o_\p (1/\sqrt{n})$ since $\frac{1}{m}\sum_{j=1}^{m}\epsilon_j$ converges to zero in probability and $\sqrt{n}(\bar{\theta}-\theta_\star)$ converges in distribution by \cref{DP SGD normal nonlinear}. 
	The third term in \eqref{eq decomposition bootstrap} is of order $o_\p(1/\sqrt{n})$ as well by \cref{convergence in prop}. Therefore
	\[\sqrt{n}\bar{\theta}^{\star}=-\frac{1}{\sqrt{n}}\sum_{j=1}^{n}\epsilon_{k_j}G^{-1}\xi_j + o_P(1)\overset{d}{\to} N(0,\Sigma)\]
	in probability, given $\theta_1,\ldots,\theta_n$ by \cref{convergence in distribution - new}(note that the sigma algebras generated by $\xi_1,\ldots,\xi_n$ and $\theta_1,\ldots,\theta_n$ coincide).

	\newpage
	
	\section{Proofs of Lemma  \ref{decomposition bootstrap} - Lemma \ref{convergence in prop} }
	
	\subsection{Proof of Lemma \ref{decomposition bootstrap}}
	Obviously,
	\begin{align*}
		\bar{\theta}^{\star}=& \frac{1}{m}\sum_{j=1}^m\epsilon_j\frac{1}l{}\sum_{b=(j-1)l+1}^{jl}(\theta_b-\theta_\star +\theta_\star- \bar{\theta})
		=\frac{1}{ml}\sum_{j=1}^m\epsilon_j\sum_{b=(j-1)l+1}^{jl}(\theta_b-\theta_\star)-(\bar\theta-\theta_\star)\frac{1}{m}\sum_{j=1}^m\epsilon_j.
	\end{align*}
	Now we use the following representation for $i$-th iterate
	\begin{equation*}
		\theta_i-\theta_\star=\prod_{j=1}^{i}(I-\eta_jG)(\theta_0-\theta_\star)-\sum_{j=1}^{i}\eta_j\xi_j\prod_{k=j+1}^{i}(I-\eta_kG)
	\end{equation*}
	to obtain
	\begin{align*}
		\bar{\theta}^\star&=\frac{1}{n}\sum_{i=1}^n\epsilon_{k_i}\left( \prod_{j=1}^{i}(I-\eta_jG)(\theta_0-\theta_\star)-\sum_{j=1}^{i}\eta_j\xi_j\prod_{k=j+1}^{i}(I-\eta_kG) \right)
		-(\bar\theta-\theta_\star)\frac{1}{m}\sum_{j=1}^m\epsilon_j\\
		&=\frac{1}{n}\sum_{i=1}^n\epsilon_{k_i} \prod_{j=1}^{i}(I-\eta_jG)(\theta_0-\theta_\star) 
		-\frac{1}{n}\sum_{j=1}^n\eta_j\xi_j\left(\sum_{i=j}^n\epsilon_{k_i}\prod_{k=j+1}^i(I-\eta_kG)\right)
		-(\bar\theta-\theta_\star)\frac{1}{m}\sum_{j=1}^m\epsilon_j\\
		&=(\theta_0-\theta_\star)\frac{1}{n \eta_1}\beta_1^n
		- \frac{1}{n}\sum_{j=1}^n \epsilon_{k_j}G^{-1}\xi_j
		- \frac{1}{n}\sum_{j=1}^n(\beta_j^n-\epsilon_{k_j}G^{-1})\xi_j
		- (\bar\theta-\theta_\star)\frac{1}{m}\sum_{j=1}^m\epsilon_j,
	\end{align*}
	which proves the assertion of \cref{decomposition bootstrap}.
	
	\subsection{Proof of Lemma \ref{convergence in distribution - new}}
	We will give the proof for $d=1$, the multivariate case follows by analog arguments. Define 
	\begin{align*}
		Y_{n,i}=\frac{1}{\sqrt{l}} \sum_{j=(i-1)l+1}^{il} \xi_j
	\end{align*}
	and note that
	\[
	\frac{1}{\sqrt{n}}\sum_{j=1}^{n}\epsilon_{k_j}G^{-1}\xi_j  =  \frac{1}{\sqrt{m}}\sum_{i=1}^m \epsilon_i G^{-1} Y_{n,i}.
	\]
	As, conditional on  
	$\xi_1,\ldots,\xi_n$, the random variables $\epsilon_1Y_{n,1},\ldots,\epsilon_mY_{n,m}$ are independent we obtain  by the Berry Esseen theorem that 
	\begin{align}
		\label{det101}
		\sup_{x\in \mathbb{R}} \Big | \p \Big (
		\frac{1}{\sqrt{m}}\sum_{i=1}^m \epsilon_i G^{-1} Y_{n,i} \leq x | \xi_1,\ldots,\xi_n
		\Big ) - \Phi \Big (\frac{ x}{\sigma_n} \Big ) \Big | \leq \frac{1}{\sqrt{m}}\frac{\frac{1}{m} \sum_{i=1}^m |G^{-1} Y_{n,i}|^3 \E[|\epsilon_i|^3]}{ \sigma_n^3}
	\end{align}
	where 
	\[
	\sigma_n^2= 
	\frac{1}{m}\sum_{i=1}^m (G^{-1}
	Y_{n,i})^2
	= 
	G^{-1} \frac{1}{m}\sum_{i=1}^m Y_{n,i}^2 G^{-1}
	\]
	is the variance of $  \frac{1}{\sqrt{m}}\sum_{i=1}^m \epsilon_i G^{-1} Y_{n,i}$
	conditional on $\xi_1, , \ldots , \xi_n$. 
	Note that, by \cref{conditions LR} (1)
	\[
	\E \Big [ \frac{1}{m} \sum_{i=1}^m Y_{n,i}^2 \Big ]  = \frac{1}{n}  \sum_{i=1}^n \E [ \xi_i^2] = S+ \frac{1}{n}  \sum_{i=1}^n 
	\E[s(\theta_{i-1}^{LDP}-\theta_\star)] = S + o(1).
	\]
	Moreover, as
	\[
	E[Y_{n,i}^4]=\frac{1}{l^2}\sum_{j=(i-1)l+1}^{il} \E[\xi_j^4]+\frac{1}{l^2}\sum_{k\neq j=(i-1)l+1}^{il}\E[\xi_j\xi_k^3]+\frac{1}{l^2}\sum_{k\neq j=(i-1)l+1}^{il}\E[\xi_j^2\xi_k^2]
	\]
	it follows that the $E[Y_{n,i}^4]$ are uniformly bounded (note that 
	that the fourth moment of $\xi_i$ is bounded by assumption). Therefore, by 
	Chebyshev's inequality, 
	$\frac{1}{m}\sum_{i=1}^m Y_{n,i}^2\overset{\p}{\rightarrow}S$, which yields
	\[
	\sigma_n^2  \overset{\p}{\rightarrow} 
	\sigma^2= G^{-1} S G^{-1}.
	\]
	By assumption, $\E[|\epsilon_i|^3]$ is bounded (since $|\epsilon_i|$ is bounded almost surely) and $\frac{1}{m} \sum_{i=1}^m |Y_{n,i}|^3$ converges in probability. Therefore, the right hand side of \eqref{det101} is of order $O_{\mathbb{P}}(1/\sqrt{m} )$ and 
	the assertion of the Lemma follows.

	\subsection{Proof of Lemma \ref{convergence in prop} }
	
	We first derive an alternative representation for the expression in \cref{convergence in prop}.
	By interchanging the order of summation we obtain
	\begin{align*}
		\frac{1}{\sqrt{n}}\sum_{i=1}^{n}(\beta_i^n-\epsilon_{k_i}G^{-1})\xi_i
		=& \frac{1}{\sqrt{n}}\sum_{i=1}^{n}\Big(\eta_i\sum_{j=i}^{n}\epsilon_{k_j}\prod_{k=i+1}^j(I-\eta_kG) - \epsilon_{k_i}G^{-1}\Big)\xi_i =
		\frac{1}{\sqrt{m}}\sum_{i=1}^{m} \epsilon_iG^{-1}V_i
	\end{align*}
	where $V_i$ are defined by
	\begin{align*}
		V_i=&\frac{1}{\sqrt{l}}\sum_{b=(i-1)l+1}^{il}\left(\sum_{j=1}^{b}G\eta_j\prod_{k=j+1}^{b}(I-\eta_kG)\xi_j-\xi_b\right).
	\end{align*}
	With these notations the assertion of \cref{convergence in prop} follows from the $L_2$ convergence
	\begin{equation*}
		\E\left[\left\lVert\frac{1}{\sqrt{n}}\sum_{i=1}^{n}(\beta_i^n-\epsilon_{k_i}G^{-1})\xi_i\right\rVert^2\right] \leq \lVert G^{-1}\rVert^2 \E\left[\left\lVert\frac{1}{\sqrt{m}}\sum_{i=1}^{m}\epsilon_{i}V_i\right\rVert^2\right] \rightarrow 0.
	\end{equation*}
	The expression on the right hand side can be decomposed into different parts according to Lemma \ref{E2 decomposition}.
	
	\begin{lemma} \label{E2 decomposition}
		Define $\Delta_i=\theta_i-\theta_\star$, $\Sigma_i=E[s(\Delta_{i-1})] $, where $s:\mathbb{R}^d\to\mathbb{R}^{d\times d}$ is the function in \cref{conditions LR}, and $\E[\xi_i\xi_i^T]=S+\Sigma_i$. It holds that
		\begin{align*}
			\E\left[\left\lVert\frac{1}{\sqrt{m}}\sum_{i=1}^{m}\epsilon_{i}V_i\right\rVert^2\right]
			=& tr\left(S \{I- R_{l,m} - N_{l,m} + NC_{l,m}\}\right) + tr(I^{rest} - R_{l,m}^{rest} - N_{l,m}^{rest} + NC_{l,m}^{rest})
		\end{align*}
		where
		\begin{align}
			R_{l,m}=&\frac{1}{lm}\sum_{i=1}^{m}\left\{\sum_{b=(i-1)l+1}^{il} \sum_{j=1}^bK_{b,j}^TK_{b,j}\right\} \nonumber\\
			N_{l,m}=&\frac{1}{lm}\sum_{i=1}^{m}\left\{
			\sum_{b=(i-1)l+1}^{il}\sum_{b'=(i-1)l+1}^{b}K_{b,b'}^T
			+\sum_{b'=(i-1)l+1}^{il}\sum_{b=(i-1)l+1}^{b}K_{b',b}\right\} \label{something negative}\\
			NC_{l,m}=&\frac{1}{lm}\sum_{i=1}^{m}\left\{
			\sum_{b=(i-1)l+1}^{il}\sum_{b'=(i-1)l+1}^{b}\sum_{j=1}^{b'}K_{b,j}^T K_{b',j} + \sum_{b'=(i-1)l+1}^{il}\sum_{b=(i-1)l+1}^{b'}\sum_{j=1}^{b}K_{b,j}^T K_{b',j}
			\right\} \nonumber \\ 
			I^{rest}=&\frac{1}{n}\sum_{i=1}^n \Sigma_i\nonumber\\
			R_{l,m}^{rest}=&\frac{1}{lm}\sum_{i=1}^{m} \sum_{b=(i-1)l+1}^{il} \sum_{j=1}^b \Sigma_j K_{b,j}^T K_{b,j} \nonumber \\ 
			N_{l,m}^{rest}=&\frac{1}{lm}\sum_{i=1}^{m}
			\sum_{b=(i-1)l+1}^{il}\sum_{b'=(i-1)l+1}^{b}\Sigma_{b'}K_{b,b'}^T
			+\sum_{b'=(i-1)l+1}^{il}\sum_{b=(i-1)l+1}^{b}\Sigma_{b} K_{b',b}\label{something negative rest}\\
			NC_{l,m}^{rest}=&\frac{1}{lm}\sum_{i=1}^{m} 
			\sum_{b=(i-1)l+1}^{il}\sum_{b'=(i-1)l+1}^{b}\left\{\sum_{j=1}^{b'}\Sigma_j K_{b,j}^T K_{b',j}\right\} 
			+ \sum_{b'=(i-1)l+1}^{il}\sum_{b=(i-1)l+1}^{b'}\left\{\sum_{j=1}^{b}\Sigma_j K_{b,j }^T K_{b',j}
			\right\} \nonumber 
		\end{align}
		
		and
		\begin{align}\label{Ks}
			K_{b,j}&:=\eta_jG\prod_{k=j+1}^{b}(I-\eta_kG).
		\end{align}
	\end{lemma}
	
	In order to estimate the expressions in \cref{E2 decomposition} we now argue as follows. For the first term we use the Cauchy-Schwarz inequality and obtain
	\[\{tr\left(S \{I- R_{l,m} - N_{l,m} + NC_{l,m}\}\right)\}^2\leq tr(S^2) tr(\{I- R_{l,m} - N_{l,m} + NC_{l,m}\}^2).\]
	Note that the matrix $I- R_{l,m} - N_{l,m} + NC_{l,m}$ is symmetric as it is a polynomial of the symmetric matrix $G$.
	Consequently, the second term converges to $0$ if we can show that all eigenvalues of the matrix $I- R_{l,m} - N_{l,m} + NC_{l,m}$ converge to zero, that is
	\[\lambda(I- R_{l,m} - N_{l,m} + NC_{l,m}) \to 0,\]
	where $\lambda(A)$ denotes the eigenvalue of the matrix $A$. For this purpose we note again that the matrix is a polynomial of the symmetric matrix $G$ which implies that 
	\begin{equation*}
		\lambda(I- R_{l,m} - N_{l,m} + NC_{l,m}) = 1- \lambda(R_{l,m}) - \lambda(N_{l,m}) + \lambda(NC_{l,m})
	\end{equation*}
	and investigate the eigenvalues of the different matrices separately. In particular, we show in Section \ref{proof1} the following results.
	
	\begin{lemma}  \label{neg 0}
		If $n=lm\to\infty$, it holds that
		\[\lambda(R_{l,m}) \to 0.\]
	\end{lemma}
	
	\begin{lemma}\label{neg 2}
		If $\frac{m^\gamma}{l^{1-\gamma}} \to 0$ and $\eta_i=ci^{-\gamma}$ for $c>0$ and $\gamma\in (0.5,1)$, it holds that 
		\[\lambda(N_{l,m})
		\rightarrow 2.\]
	\end{lemma} 
	
	\begin{lemma} \label{pos 1} 
		If $\frac{m^\gamma}{l^{1-\gamma}} \to 0$ and $\log(l)/m \to 0$, it holds that
		\[\lambda(NC_{l,m})
		\rightarrow 1.\]   
	\end{lemma}
	
	\noindent From these results we can conclude that 
	\begin{equation} \label{trace to zero}
		tr(S\{I- R_{l,m} - N_{l,m} + NC_{l,m}\}) \to 0.
	\end{equation}
	In order to derive a similar result for the second term in \cref{E2 decomposition} we consider all four terms in the trace separately.
	Starting with $I^{Rest}$ we obtain by \cref{conditions LR} that
	\[|tr(I^{rest})| \leq \frac{1}{n}\sum_{i=1}^n \sqrt{tr(\Sigma_i^2)} \leq \frac{1}{n}\sum_{i=1}^n \sqrt{\eta_i} \to 0 .\]
	Next we note, observing the definition of $R_{l,m}^{rest}$ that
	\[|tr(R_{l,m}^{rest})|\leq \frac{1}{lm}\sum_{i=1}^{m} \sum_{b=(i-1)l+1}^{il} \sum_{j=1}^b \sqrt{tr(\Sigma_j^2)} tr(K_{b,j}^T K_{b,j}) \leq C tr(R_{l,m}), \]
	since $\sqrt{tr(\Sigma_j^2)}$ can be bounded by a constant by \cref{conditions LR}. By \cref{neg 0} $tr(R_{l,m})$ converges to zero which implies that 
	\[tr(R_{l,m}^{rest}) \to 0.\]
	The two remaining terms $tr(N_{l,m}^{rest})$ and $tr(NC_{l,m}^{rest})$ are considered in the following Lemma, which will be proved in Section \ref{proof1}.
	\begin{lemma}\label{neg 2 rest}
		Under \cref{conditions LR} it holds that 
		\begin{equation} \label{neg 2 rest 1}
			tr(N_{l,m}^{rest})\rightarrow 0
		\end{equation}
		and that
		\begin{equation} \label{neg 2 rest 2}
			tr(NC_{l,m}^{rest})\rightarrow 0.
		\end{equation}
	\end{lemma} 
	Combining these arguments yield
	\[tr(I^{rest} - R_{l,m}^{rest} - N_{l,m}^{rest} + NC_{l,m}^{rest})\to 0 \]
	and the assertion of \cref{convergence in prop} follows from \cref{E2 decomposition} and \eqref{trace to zero}.

	\newpage
	
	\section{Proofs of auxiliary results}
	
	\label{proof1}
	\subsection{Proof of \cref{E2 decomposition}}
	Observing that the bootstrap multipliers $\epsilon_i$ are independent of $\xi_j$ it follows that 
	\begin{align*}
		\E\left[\left\lVert\frac{1}{\sqrt{m}}\sum_{i=1}^{m}\epsilon_{i}V_i\right\rVert^2\right]
		&=\frac{1}{m}\sum_{i=1}^{m}\E[\epsilon_{i}^2V_i^TV_i]+\frac{1}{m}\sum_{i\neq j=1}^{m}\E[\epsilon_{j}\epsilon_{i}V_i^TV_j]
		=\frac{1}{m}\sum_{i=1}^{m}\E[V_i^TV_i]
	\end{align*}
	since $\E[\epsilon_i]=0$ and $Var(\epsilon_i)=1$.
	With the notation 
	\begin{align*}
		B_b&:=\sum_{j=1}^{b}\eta_jG\prod_{k=j+1}^{b}(I-\eta_kG)\xi_j-\xi_b=\sum_{j=1}^{b}K_{b,j}\xi_j-\xi_b,
	\end{align*}
	we obtain the representation $V_i=\frac{1}{\sqrt{l}}\sum_{b=(i-1)l+1}^{il}B_b$ which yields
	\begin{align*}
		\E[V_i^TV_i]=\frac{1}{l}\sum_{b=(i-1)l+1}^{il}\E[B_b^TB_b]+\frac{1}{l}\sum_{b\neq b'=(i-1)l+1}^{il}\E[B_b^TB_{b'}]
	\end{align*}
	where 
	\begin{align*}
		\E[B_b^TB_b]&=
		\E\left[\left(\sum_{j=1}^{b}K_{b,j}\xi_j\right)^T\left(\sum_{j=1}^{b}K_{b,j}\xi_j\right)\right]
		-2\sum_{j=1}^{b}\E[\xi_b^TK_{b,j}\xi_j]
		+\E[\xi_b^T\xi_b]\\
		&=\sum_{j=1}^{b}\E[\xi_j^TK_{b,j}^TK_{b,j}\xi_j] 
		- 2\E[\xi_b^TK_{b,b}\xi_b] + \E[\xi_b^T\xi_b]\\
		&=\sum_{j=1}^{b}tr(\E[\xi_j\xi_j^T]K_{b,j}^TK_{b,j}) 
		- 2tr(\E[\xi_b\xi_b^T]K_{b,b}) + tr(\E[\xi_b\xi_b^T]),
	\end{align*}
	since $\E[\xi_i^TA\xi_j]=\E[\E[\xi_i^TA\xi_j|\mathcal{F}_{max\{i,j\}-1}]]=0$ for $i\neq j$.
	For $b\neq b'$ a similar calculation shows that 
	\begin{align*}
		\E[B_b^TB_{b'}]&=\E\left[\left(\sum_{j=1}^{b}\xi_jK_{b,j}-\xi_b\right)^T\left(\sum_{i=1}^{b'}\xi_iK_{b',i}-\xi_{b'}\right)\right]\\
		&=\sum_{j=1}^{min(b,b')}tr(\E[\xi_j\xi_j^T]K_{b,j}^TK_{b',j}) - \mathbb{I}\{b<b'\} tr(\E[\xi_b\xi_b^T]K_{b',b}) - \mathbb{I}\{b'<b\} tr(\E[\xi_{b'}\xi_{b'}^T]K_{b,b'}^T).
	\end{align*}
	Inserting $\E[\xi_j\xi_j^T]= S+\Sigma_j$ and noting that $K_{b,b}=\eta_bG$ it follows
	\begin{align*}
		\E[V_i^TV_i]=&\frac{1}{l}\sum_{b=(i-1)l+1}^{il}\left\{
		\sum_{j=1}^{b}tr((S+\Sigma_j) K_{b,j}^TK_{b,j}) 
		- 2tr((S+\Sigma_b) K_{b,b}) + tr(S+\Sigma_b) \right\}\\
		&+\frac{1}{l}\sum_{b=(i-1)l+1}^{il}\sum_{b\neq b'=(i-1)l+1}^{il} \left\{ \sum_{j=1}^{min(b,b')}tr((S+\Sigma_j) K_{b,j}^TK_{b',j}) \right\}\\
		&- \frac{1}{l}\sum_{b=(i-1)l+1}^{il}\sum_{b\neq b'=(i-1)l+1}^{il} \left\{
		\mathbb{I}\{b<b'\} tr((S+\Sigma_b) K_{b',b}) + \mathbb{I}\{b'<b\} tr((S+\Sigma_{b'}) K_{b,b'}^T) \right\}
	\end{align*}
	The assertion of \cref{E2 decomposition} now follows by a straight forward calculation adding and subtracting the terms with $b=b'$ in the second and third sum.

	\subsection{Proof of Lemma \ref{neg 0}}
	As $R_{l,m}$ is a polynomial of the symmetric matrix $G$ we may assume without loss of generality that $\lambda(G)=1$ (change the constant $c$ in the definition of $\eta_k$) and obtain
	with the inequalities $1+x\leq e^x$ and $\sum_{i=1}^k \eta_i\geq c\int_1^{k+1}x^{-\gamma}dx$ that
	\begin{align*}
		\lambda(R_{l,m})
		=&\frac{1}{n}\sum_{i=1}^{n} \left[\sum_{j=1}^i\eta_j^2\prod_{k=j+1}^i(1-\eta_k)^2\right]
		\leq  \frac{1}{n}\sum_{i=1}^{n} \left[\sum_{j=1}^i\eta_j^2\exp\left(-2\sum_{k=j+1}^i\eta_k\right)\right]\\
		\leq & \frac{1}{n}\sum_{i=1}^{n} \left[\sum_{j=1}^i\eta_j^2\exp\left(-2c\int_{j+1}^{i+1}x^{-\gamma}dx\right)\right]\\
		=&\frac{1}{n}\sum_{i=1}^{n} \exp\left(-\tfrac{2c}{1-\gamma}(i+1)^{1-\gamma}\right) \left[\sum_{j=1}^i\eta_j^2\exp\left(\tfrac{2c}{1-\gamma}(j+1)^{1-\gamma}\right)\right]
	\end{align*}
	Fixing a constant $v\in (0.5,1)$, the inner sum can be split at $\lfloor iv \rfloor$, i.e.
	\[\lambda(R_{l,m}) \leq \frac{1}{n}\sum_{i=1}^{n} \exp\left(-\tfrac{2c}{1-\gamma}(i+1)^{1-\gamma}\right) \left[F_i+L_i\right],\]
	where $F_i=\sum_{j=1}^{\lfloor vi \rfloor}\eta_j^2\exp\left(\frac{2c}{1-\gamma}(j+1)^{1-\gamma}\right)$ and $L_i=\sum_{j=\lfloor vi \rfloor+1}^{i}\eta_j^2\exp\left(\frac{2c}{1-\gamma}(j+1)^{1-\gamma}\right)$.
	It holds that 
	\begin{align*}
		F_i &\leq \exp\left(\tfrac{2c}{1-\gamma}(\lfloor vi \rfloor+1)^{1-\gamma}\right) \sum_{j=1}^{\lfloor vi \rfloor}\eta_j^2
		\leq c^2\exp\left(\tfrac{2c}{1-\gamma}(\lfloor vi \rfloor+1)^{1-\gamma}\right)\left(1+\int_1^{\lfloor vi \rfloor} x^{-2\gamma} dx\right)\\
		L_i & \leq \exp\left(\tfrac{2c}{1-\gamma}((i+1)^{1-\gamma}\right)\sum_{j=\lfloor vi \rfloor+1}^{i}\eta_j^2 \leq c^2 \exp\left(\tfrac{2c}{1-\gamma}((i+1)^{1-\gamma}\right )\int_{\lfloor vi \rfloor}^i x^{-2\gamma} dx
	\end{align*}
	which yields
	\begin{align*}
		\lambda(R_{l,m}) \leq &
		\frac{c^2}{n}\sum_{i=1}^n \exp\left(-\tfrac{2c}{1-\gamma}(i+1)^{1-\gamma} (1-(\tfrac{\lfloor v i \rfloor+1}{i+1})^{1-\gamma})\right)
		(1+\tfrac{1-\lfloor v i \rfloor^{1-2\gamma}}{2\gamma -1})\\
		& + \frac{c^2}{n}\sum_{i=1}^ni^{1-2\gamma} \frac{1}{2\gamma-1}((\tfrac{i}{\lfloor vi \rfloor})^{2\gamma-1} -1)
	\end{align*}
	
	It is easy to see that $(\lfloor vi \rfloor+1)/(i+1)\leq (v+1)/2$ and $\tfrac{i}{\lfloor vi \rfloor}\leq \tfrac{1}{v-1/2}$ for $i\geq 2$. Therefore and since $\lfloor vi \rfloor^{1-2\gamma}\geq 0$,
	
	\begin{align*}
		\lambda(R_{l,m}) \leq & \tfrac{2\gamma c^2}{(2\gamma-1)n}\sum_{i=1}^{n} \exp\left(-\tfrac{2c}{1-\gamma}(i+1)^{1-\gamma}\left(1-\left(\tfrac{v+1}{2}\right)^{1-\gamma}\right)\right)  + \tfrac{\kappa c^2}{n(2\gamma-1)}\sum_{i=1}^{n} i^{1-2\gamma},
	\end{align*} 
	where $\kappa=(\tfrac{1}{v-1/2})^{2\gamma -1} -1>0$.
	Since  $\frac{1}{n}\sum_{i=1}^n\exp (-K(i+1)^{1-\gamma})\rightarrow 0$ for $K>0$ and $\frac{1}{n}\sum_{i=1}^{n} i^{1-2\gamma}\to 0$, the assertion of the lemma follows.

	\subsection{Proof of Lemma \ref{neg 2}}
	Without loss of generality we assume that $\lambda(G)=1$ (changing the constant $c$ in the learning rate). Because $N_{l,m}$ is a polynomial of the symmetric matrix G it follows that 
	\[\lambda(N_{l,m})=\frac{2}{lm}\sum_{i=1}^{m}\sum_{b=(i-1)l+1}^{il}\sum_{b'=(i-1)l+1}^{b}\left\{\eta_{b'}\prod_{k=b'+1}^b(1-\eta_k)\right\}.\]
	Note that 
	\begin{equation}\label{value of sum}
		\sum_{j=1}^b \eta_j \prod_{k=j+1}^b (1-\eta_k) = 1- \prod_{k=1}^b (1-\eta_k)
	\end{equation}
	which follows by a direct calculation using an induction argument.
	With this representation we obtain
	\begin{align*}
		\sum_{b'=(i-1)l+1}^{b}\left\{\eta_{b'}\prod_{k=b'+1}^b(1-\eta_k)\right\} 
		= & \sum_{b'=1}^{b}\left\{\eta_{b'}\prod_{k=b'+1}^b(1-\eta_k)\right\}-\sum_{b'=1}^{(i-1)l}\left\{\eta_{b'}\prod_{k=b'+1}^b(1-\eta_k)\right\} \\
		=&  1- \prod_{k=1}^b (1-\eta_k)-\left( 1-\prod_{k=1}^{(i-1)l} (1-\eta_k)\right)\prod_{k=(i-1)l+1}^b(1-\eta_k)\\
		=& 1-  \prod_{k=(i-1)l+1}^b(1-\eta_k).
	\end{align*}
	Therefore, it is left to show that 
	\[A_n:=
	\frac{2}{lm}\sum_{i=1}^{m}\sum_{b=(i-1)l+1}^{il} \prod_{k=(i-1)l+1}^b(1-\eta_k) \rightarrow 0.
	\]
	For $k\leq b\leq il$ use $(1-\eta_k)\leq (1-\eta_{il})$ which gives (using the definition $\eta_k=ck^{-\gamma}$)
	\begin{align*}
		\sum_{b=(i-1)l+1}^{il} \prod_{k=(i-1)l+1}^b(1-\eta_k)
		\leq & \sum_{b=(i-1)l+1}^{il}  (1-\eta_{il})^{b-(i-1)l}
		= (1-\eta_{il}) \frac{1-(1-\eta_{il})^l}{\eta_{il}}\\
		= & (i^\gamma l^\gamma -c)\frac{1-(1-ci^{-\gamma} l^{-\gamma})^l}{c}
		\leq \frac{1}{c}i^\gamma l^\gamma
	\end{align*}
	if $l$ is sufficiently large. This gives
	\begin{align*}
		A_n
		\leq & \frac{2}{lmc}\sum_{i=1}^{m} i^\gamma l^\gamma \leq \frac{2m^\gamma}{cl^{1-\gamma}},
	\end{align*}
	which converges to zero since $\frac{m^\gamma}{l^{1-\gamma}} \to 0$, by assumption.

	\subsection{Proof of Lemma \ref{pos 1}}
	As in the proof of \cref{pos 1} we assume  without loss of generality that $\lambda(G)=1$. Observing again that $NC_{l,m}$ is a polynomial of the symmetric matrix $G$ it follows that 
	\[\lambda(NC_{l,m})=\frac{2}{lm}\sum_{i=1}^{m}\sum_{b=(i-1)l+1}^{il}\sum_{b'=(i-1)l+1}^{b-1}\left\{\sum_{j=1}^{b'}\eta_j^2\prod_{k=j+1}^{b'}(1-\eta_k)^2\right\}\prod_{k=b'+1}^b(1-\eta_k).\]
	We will show below that there exists a constant $c'$ and a constant $b_0$  (which depends on the parameters of the learning rate) such that for all $b\geq b_0$
	\begin{align}
		\sum_{j=1}^{b}\eta_j^2\prod_{k=j+1}^{b}(1-\eta_k)^2 & \leq \eta_b/2 + c' b^{-1} \label{OC upper bound} \\
		\sum_{j=1}^{b}\eta_j^2\prod_{k=j+1}^{b}(1-\eta_k)^2 & \geq \eta_b/2 \label{OC lower bound}.
	\end{align}
	Recalling the definition of $N_{l,m}$ in \eqref{something negative} and using \eqref{OC upper bound}, \eqref{OC lower bound}  and \cref{neg 2} it follows that 
	\begin{align}
		\lambda(NC_{l,m})&\leq \lambda(N_{l,m})/2 +c'\frac{2}{lm}\sum_{i=1}^{m}\sum_{b=(i-1)l+1}^{il}\sum_{b'=(i-1)l+1}^{b-1}b'^{-1} \to 1, \label{limit smaller 1} \\
		\lambda(NC_{l,m})&\geq \lambda(N_{l,m})/2\rightarrow 1  \nonumber
	\end{align}
	since the first term in \eqref{limit smaller 1} converges to $1$ by \cref{pos 1} while the second term vanishes asymptotically, which is a consequence of the following
	\begin{align*}
		\frac{2}{lm}\sum_{i=1}^{m}\sum_{b=(i-1)l+1}^{il}\sum_{b'=(i-1)l+1}^{b}b'^{-1}
		& = \frac{2}{lm}\sum_{i=1}^{m}\sum_{b'=(i-1)l+1}^{il} \sum_{b=b'+1}^{il }b'^{-1}\\
		& = \frac{2}{lm}\sum_{i=1}^{m}\sum_{b'=(i-1)l+1}^{il} b'^{-1}(il-b')\\
		& = \frac{2}{m}\sum_{i=1}^{m}i\sum_{b'=(i-1)l+1}^{il}b'^{-1} -2\\
		& \leq \frac{2}{m}\sum_{b'=1}^{l}b'^{-1} + \frac{2}{m}\sum_{i=2}^{m}i \log \left(\frac{i}{i-1}\right) -2 + o(1)\\
		&\leq \frac{2\log(l)}{m} + \frac{2}{m}\int_{2}^{m+1}x \log \left(\frac{x}{x-1}\right)dx  -2 + o(1).
	\end{align*}
	The right hand side converges to 0 since $\frac{\log(l)}{m}\to 0$ and 
	\[\frac{2}{m}\int_{2}^{m+1}x \log \left(\frac{x}{x-1}\right)dx \to 2.\]
	\bigskip\\
	Therefore it remains to prove the two inequalities \eqref{OC upper bound} and \eqref{OC lower bound}.
	\begin{proof}[Proof of \eqref{OC upper bound}]
		We will show this result by induction over $b$. Denote
		\[B= B(c')=
		\max \left\{ \left(3\frac{c^2}{4c'-1}\right)^{\frac{1}{2\gamma-1}},
		\left(3\frac{2c'}{c(4c'-1)}\right)^{\frac{1}{1-\gamma}},
		\left(3\frac{2cc'}{4c'-1}\right)^{\frac{1}{\gamma}}
		\right\} . \]
		At first we argue that there exist  constants $c'\geq 1$ and $b_0\geq B$ such that \eqref{OC upper bound} holds for $b=b_0$  (this particular choice of $B$ is used in the induction step).\\
		To see this note that, for $c' \geq 1$, $B(c')$ is bounded by a constant (depending on $c$ and $\gamma$). 
		Let $b_0$ be fixed and larger than this constant. In particular, the choice of $b_0$ does not depend on $c'$ and $b_0\geq B(c')$ for all $c'\geq 1$.
		Then, for $c'$ sufficiently large, \eqref{OC upper bound} holds for $b=b_0$ since the left-hand side is finite for $b_0$ fixed.\\
		For the induction step we assume that \eqref{OC upper bound} holds for some $b-1\geq b_0$, then we have to show:
		\begin{align} \label{induction}
			\sum_{j=1}^{b}\eta_j^2 \prod_{k=j+1}^{b}(1-\eta_k)^2 
			=&\eta_b^2 + (1-\eta_b)^2 \sum_{j=1}^{b-1}\eta_j^2\prod_{k=j+1}^{b-1}(1-\eta_k)^2 
			\leq \eta_b/2 + c' \frac{1}{b}.
		\end{align}
		Since we assumed that \eqref{OC upper bound} holds for $b-1$, \eqref{induction} follows from
		\[\eta_b^2 + (1-\eta_b)^2 \left(\eta_{b-1}/2 + c' \frac{1}{b-1} \right)
		\leq \eta_b/2 + c' \frac{1}{b}\]
		which is equivalent to (inserting the definition of $\eta_b$ and multiplying by $b^\gamma$ 
		)
		\begin{align} \label{ineq}
			\left(\left(\tfrac{b}{b-1}\right)^\gamma - 1\right)\left(-c^2b^{-\gamma}  +\tfrac{c}{2}\right) +\tfrac{1}{2}c^3b^{-2\gamma} \left(\tfrac{b}{b-1}\right)^{a}+\tfrac{c'c^2}{b-1}b^{-\gamma} + \tfrac{c'b^\gamma}{b(b-1)} \leq \tfrac{2cc'}{b-1}.
		\end{align} 
		Since $(\frac{b}{b-1})^\gamma \leq \frac{b}{b-1}$ it holds that 
		\begin{align*}
			& \left(\left(\tfrac{b}{b-1}\right)^\gamma - 1\right)\left(-c^2b^{-\gamma}  +\tfrac{1}{2}c\right) 
			+\tfrac{1}{2}c^3b^{-2\gamma} \left(\tfrac{b}{b-1}\right)^{a}+\tfrac{c'c^2}{b-1}b^{-\gamma} + \tfrac{c'b^\gamma}{b(b-1)}\\
			\leq & \tfrac{c}{2}\left(\tfrac{b}{b-1} - 1\right) +\tfrac{c^3}{2(b-1)}b^{1-2\gamma}+\tfrac{c'c^2}{b-1}b^{-\gamma} +\tfrac{c'b^\gamma}{b(b-1)}\\
			= & \frac{c +c^3b^{1-2\gamma} +2c'c^2 b^{-\gamma} +2c'b^{\gamma-1}}{2(b-1)}.
		\end{align*}
		So $\eqref{ineq}$ and therefore \eqref{induction} follows from
		\begin{align*}
			c^3b^{1-2\gamma} +2c'c^2 b^{-\gamma} +2c'b^{\gamma-1} \leq  4cc'-c,
		\end{align*}
		which is implied if the following three inequalities hold
		\begin{align*}
			c^3b^{1-2\gamma}  &\leq \frac{1}{3} c(4c'-1)\\
			2c'c^2 b^{-\gamma}  &\leq \frac{1}{3} c(4c'-1)\\
			2c'b^{\gamma-1} &\leq \frac{1}{3} c(4c'-1)
		\end{align*}
		which they do for $b\geq B$.
	\end{proof}
	\begin{proof}[Proof of \eqref{OC lower bound}]
		We will show this result by induction over $b$. By \cref{conditions LR}  there exists a constant $b_0\geq (2c)^\frac{1}{\gamma}$ such that \eqref{OC lower bound} holds for a $b-1\geq b_0$. Therefore,
		\begin{align*}
			\sum_{j=1}^{b}\eta_j^2\prod_{k=j+1}^{b}(1-\eta_k)^2 =& \eta_b^2 +(1-\eta_b)^2 \sum_{j=1}^{b-1}\eta_j^2\prod_{k=j+1}^{b-1}(1-\eta_k)^2
			\geq  \eta_b^2 +(1-\eta_b)^2 \frac{\eta_{b-1}}{2}
		\end{align*}
		by the induction hypothesis. Showing that this is larger than $\eta_b/2$ is equivalent to (inserting the definition of $\eta_b$, multiplying by $2b^{2\gamma}(b-1)^{\gamma}/c$)
		\begin{align*}
			2c\left((b-1)^\gamma -b^\gamma\right) +c^2 +b^{2\gamma} &\geq (b-1)^\gamma b^\gamma.
		\end{align*}
		Since $c^2>0$, this is implied by
		\begin{align*}
			2c &\leq b^\gamma
		\end{align*}
		(note that $(b-1)^\gamma -b^{\gamma}\leq 0$). By \cref{conditions LR} (5) we have $b^\gamma\geq b_0^\gamma\geq 2c$, which completes the proof of \cref{pos 1}.
	\end{proof}
	\subsection{Proof of Lemma \ref{neg 2 rest}}
	We start by showing \eqref{neg 2 rest 1}. 
	Note that by definition of $N_{l,m}^{rest}$ \eqref{something negative rest}
	\[tr(N_{l,m}^{rest})=\frac{2}{lm}\sum_{i=1}^{m}\sum_{b=(i-1)l+1}^{il}\sum_{b'=(i-1)l+1}^{b}tr(\Sigma_{b'}K_{b,b'})\]
	where the matrix $K_{b,b'}$ is defined in \eqref{Ks} and $\Sigma_b$ as in \cref{E2 decomposition}.
	From the Cauchy-Schwarz inequality it follows that $|tr(\Sigma_{b'}K_{b,b'}^T)|\leq \sqrt{tr(\Sigma_{b'}^2)tr(K_{b,b'}^2)}$.  
	By \cref{conditions LR}, $\sqrt{tr(\Sigma_{b'}^2)}$ is (up to a constant) bounded by $ \sqrt{\eta_{b'}}$. 
	We will prove 
	\begin{align}\label{Bn}
		B_n &:= \frac{2}{lm}\sum_{i=2}^{m}\sum_{b=(i-1)l+1}^{il}\sum_{b'=(i-1)l+1}^{b}\sqrt{\eta_{b'}\lambda(K_{b,b'}^2)}\to 0,\\
		C_n &:= \frac{2}{lm}\sum_{b=1}^{l}\sum_{b'=1}^{b}\sqrt{\eta_{b'}\lambda(K_{b,b'}^2)}\to 0 \label{Cn}
	\end{align} 
	for every eigenvalue of $K_{b,b'}$ from which \eqref{neg 2 rest 1} follows.\\
	To see \eqref{Bn}, note again that $K_{b,b'}$ is a polynomial of the symmetric matrix $G$ and as in the proof of \cref{neg 2} we assume without loss of generality that $\lambda(G)=1$, which gives
	\[\lambda(K_{b,b'}^2)=\eta_{b'}^2 \prod_{k=b'+1}^b(1-\eta_k)^2.\]
	If $l$ is sufficiently large, $1-\eta_k\geq 0$ for $k\geq b'\geq l$, and therefore 
	\begin{align*}
		\sqrt{\lambda(K_{b,b'}^2)}=\eta_{b'} \prod_{k=b'+1}^b(1-\eta_k).
	\end{align*}
	By $\sqrt{\eta_{b'}}\leq \sqrt{\eta_{(i-1)l}}$ for $b'\geq (i-1)l$ and by applying \eqref{value of sum}, it follows that 
	\begin{align*}
		B_n \leq & \frac{2}{lm}\sum_{i=2}^{m}\sqrt{\eta_{(i-1)l}}\sum_{b=(i-1)l+1}^{il}\sum_{b'=(i-1)l+1}^{b}\left\{\eta_{b'} \prod_{k=b'+1}^b(1-\eta_k)\right\}\\
		=&\frac{2}{lm}\sum_{i=2}^{m}\sqrt{\eta_{(i-1)l}}\sum_{b=(i-1)l+1}^{il}(1-\prod_{k=(i-1)l+1}^b (1-\eta_k))\\
		\leq & \frac{2}{m}\sum_{i=2}^{m}\sqrt{\eta_{(i-1)l}}\\
		\leq &  2l^{-\gamma/2} 
	\end{align*}
	which converges to zero. \\
	\eqref{Cn} follows by similar arguments and noting that $(1-\eta_k)<0$ for only finitely many $k$.\\
	\medskip
	
	To show the second assertion \eqref{neg 2 rest 2} in \cref{neg 2 rest}, note that by the definition \eqref{something negative rest} of $NC_{l,m}^{rest}$ and same arguments as in the proof of \eqref{neg 2 rest 1}, it follows that 
	\[tr(NC_{l,m})\leq \frac{2}{lm}\sum_{i=1}^{m}\sum_{b=(i-1)l+1}^{il}\sum_{b'=(i-1)l+1}^{b-1}\left\{\sum_{j=1}^{b'}\sqrt{\eta_{b'}tr((K_{b,j}^T K_{b',j})^2)}\right\}\]
	where $tr((K_{b,j}^T K_{b',j})^2)\leq d \lambda_{max}((K_{b,j}^T K_{b',j})^2)$.
	Noting $K_{b,j}$ is a polynomial of the symmetric matrix $G$ and assuming without loss of generality that for the corresponding eigenvalue of $G$ it holds $\lambda(G)=1$ it follows
	\[\sqrt{\lambda((K_{b,j}^T K_{b',j})^2))}=\eta_j^2\prod_{k=j+1}^{b'} (1-\eta_k)^2\prod_{k=b'+1}^b |1-\eta_k|.\]
	Similar to \eqref{OC upper bound} one can show by induction that ($c$ being the constant in the learning rate)
	\begin{equation*}
		\sum_{j=1}^{b'}\eta_j^2\sqrt{\eta_j}\prod_{k=j+1}^{b'}(1-\eta_k)^2\leq c\eta_{b'}\sqrt{\eta_{b'}}+ c' b'^{-1}.
	\end{equation*}
	This implies
	\[tr(NC_{l,m})\leq \frac{2\sqrt{d}}{lm}\sum_{i=1}^{m}\sum_{b=(i-1)l+1}^{il}\sum_{b'=(i-1)l+1}^{b-1}\eta_{b'}\sqrt{\eta_{b'}}\prod_{k=b'+1}^b|1-\eta_k|+ c' b'^{-1},\]
	where the sum over the first term converges to zero by the same arguments used for the convergence of $B_n$ and $C_n$ and the sum over the second term converges to zero by the same arguments used in the proof of \eqref{OC upper bound}. Therefore \eqref{neg 2 rest 2} follows.

\end{document}